\documentclass[lettersize,journal]{IEEEtran}
\usepackage{amsmath,amsfonts}
\usepackage{algorithmic}
\usepackage{algorithm}
\usepackage{array}
\usepackage[caption=false,font=normalsize,labelfont=sf,textfont=sf]{subfig}
\usepackage{textcomp}
\usepackage{stfloats}
\usepackage{url}
\usepackage{balance}
\usepackage{verbatim}
\usepackage{graphicx}
\usepackage{cite}
\usepackage{amssymb}
\usepackage[dvipsnames]{xcolor}
\usepackage{booktabs} 
\usepackage{makecell}
\usepackage{multirow}
\usepackage{enumitem}

\usepackage[utf8]{inputenc}
\usepackage[colorlinks=true,linkcolor=blue,citecolor=blue,urlcolor=blue]{hyperref}
\usepackage{bm}
\usepackage{bigstrut}
\usepackage{graphicx}

\begin{document}
	
	\title{Prototype-Driven Structure Synergy Network for Remote Sensing Images Segmentation}
	
	\author{Junyi~Wang,  Jinjiang Li, Guodong Fan, Yakun Ju, ~\IEEEmembership{Member,~IEEE}, Xiang Fang,  Alex C. Kot,~\IEEEmembership{Life Fellow,~IEEE}
        \thanks{This research was supported by the National Natural Science Foundation of China (62301105, 61772319, 62272281, 62002200, 62202268), Shandong Natural Science Foundation of China (ZR2020QF012 and ZR2021MF068), Yantai science and technology innovation development plan(2022JCYJ031).}
		\thanks{J. Wang, J. Li and G. Fan are with School of Computer Science and Technology, Shandong Technology and Business University, Yantai 264005, China. (2024410067@sdtbu.edu.cn, lijinjiang@gmail.com, fgd96@outlook.com)}
            
	\thanks{Yakun Ju is with the School of Computing and Mathematical Sciences, University of Leicester, Leicester, LE1 7RH, United Kingdom. (kelvin.yakun.ju@gmail.com)}
    \thanks{Xiang Fang is with the College of Computing and Data Science, Nanyang Technological University, 50 Nanyang Avenue, 639798, Singapore. (xfang9508@gmail.com)}
    \thanks{Alex C. Kot is with the Director of the Rapid-Rich Object Search (ROSE) Laboratory and the NTU-PKU Joint Research Institute, Nanyang Technological University, 50 Nanyang Avenue, 639798, Singapore. (eackot@ntu.edu.sg)}
	}
	
	\markboth{Journal of \LaTeX\ Class Files,~Vol.~14, No.~8, August~2021}%
	{Shell \MakeLowercase{\textit{et al.}}: A Sample Article Using IEEEtran.cls for IEEE Journals}
	
	\IEEEpubid{}
	
	\maketitle
	
	\begin{abstract}
		In the semantic segmentation of remote sensing images, acquiring complete ground objects is critical for achieving precise analysis. However, this task is severely hindered by two major challenges: high intra-class variance and high inter-class similarity. Traditional methods often yield incomplete segmentation results due to their inability to effectively unify class representations and distinguish between similar features. Even emerging class-guided approaches are limited by coarse class prototype representations and a neglect of target structural information.
		Therefore, this paper proposes a Prototype-Driven Structure Synergy Network (PDSSNet). The design of this network is based on a core concept, a complete ground object is jointly defined by its invariant class semantics and its variant spatial structure. To implement this, we have designed three key modules. First, the Adaptive Prototype Extraction Module (APEM) ensures semantic accuracy from the source by encoding the ground truth to extract unbiased class prototypes. Subsequently, the designed Semantic-Structure Coordination Module (SSCM) follows a hierarchical “semantics-first, structure-second” principle. This involves first establishing a global semantic cognition, then leveraging structural information to constrain and refine the semantic representation, thereby ensuring the integrity of class information. Finally, the Channel Similarity Adjustment Module (CSAM) employs a dynamic step-size adjustment mechanism to focus on discriminative features between classes.
		Extensive experiments demonstrate that PDSSNet outperforms state-of-the-art methods. The source code is available at https://github.com/wangjunyi-1/PDSSNet.
	\end{abstract}
	
	
	\section{INTRODUCTION}
	\IEEEPARstart{w}{ith} 
	the advancement of aerospace sensors, remote sensing images now enable fine-grained observation of the Earth's surface. Semantic segmentation, a core technology in this field, transforms pixels into structured geographical objects. Its practical applications, such as land cover mapping \cite{3,4}, change detection \cite{5,6}, and environmental protection \cite{7,8}, depend on the complete parsing of class instances. 
	
\begin{figure}[!t]
		\centering
		\includegraphics[width=3.5in]{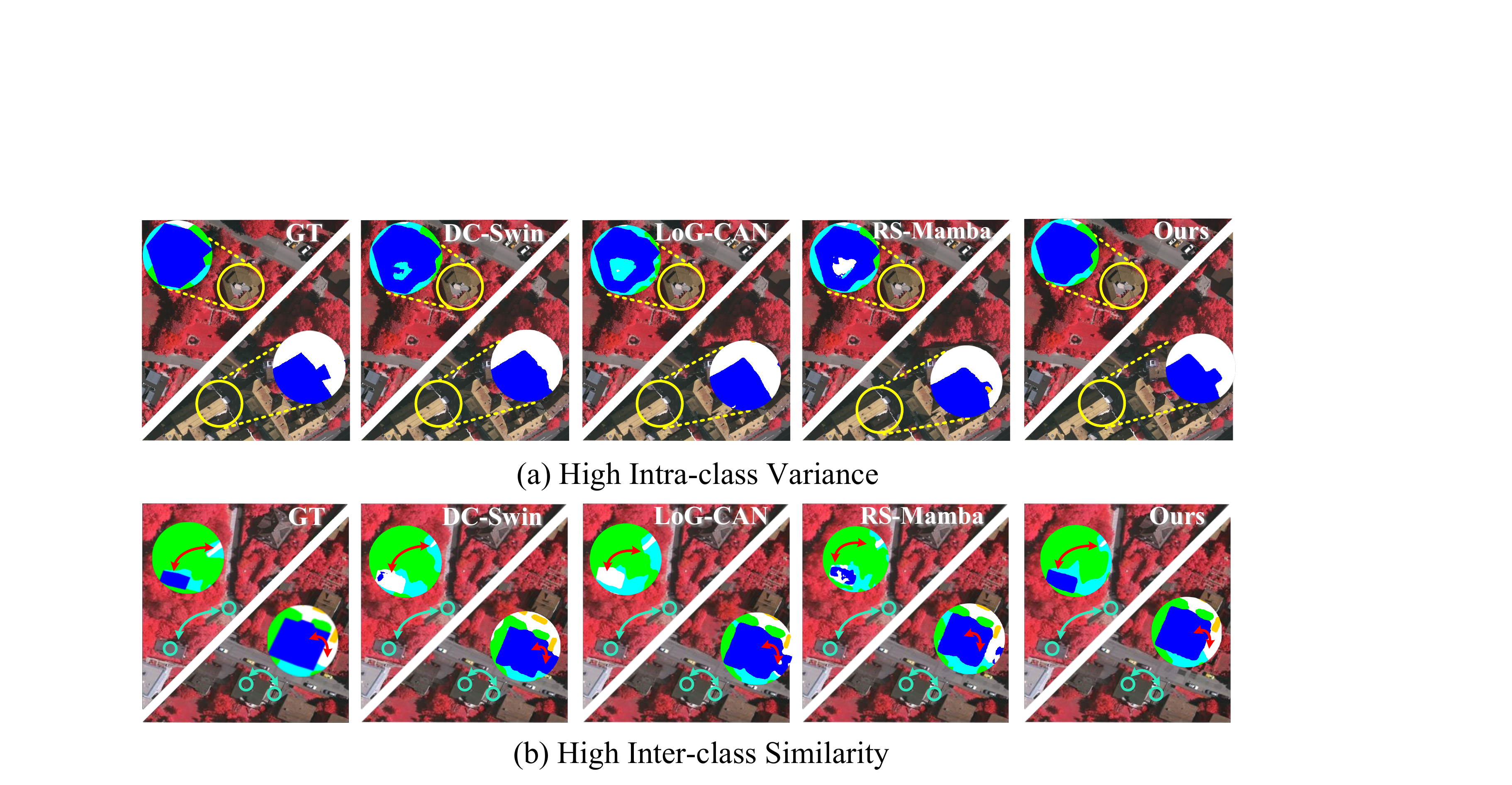}
             \vspace{-20pt}
		\caption{This figure illustrates the problem of incomplete class segmentation, where objects that should be continuous are erroneously segmented, resulting in internal holes, fragmented overall structures, and the misidentification of one class as a similar one. We find that the root cause of this incomplete segmentation can be attributed to two main challenges. The first, as shown by the yellow circle in (a), is high intra-class variance, where significant appearance differences within the same object class make it difficult for the model to learn unified features. The second, in the region indicated by the blue arrow in (b), is high inter-class similarity, where the similar features of different classes make them easy for the model to confuse.
		}
		\label{fig1}
		
	\end{figure}

	However, existing methods exhibit limitations in the completeness of their segmentation results (as illustrated in Fig. \ref{fig1}). We find that this problem primarily stems from two inherent characteristics of remote sensing images.
	On the one hand, high intra-class variance stems from the vast differences in scale, morphology, material, and lighting among objects of the same class. This resulting feature diversity directly hinders the model from learning a unified class representation capable of encompassing all variations and is a primary cause of incomplete object segmentation. On the other hand, high inter-class similarity means that different classes of ground objects exhibit similar spectral and texture features due to their comparable material compositions. This feature ambiguity makes it easy for the model to confuse different classes, thereby resulting in erroneous segmentation.

	To address the challenges posed by the aforementioned characteristics, we first analyze the current mainstream technical approaches. Models based on Convolutional Neural Networks\cite{11, 12, 15,91} and Transformers\cite{17, 18, 19,92} have achieved great success through semantic label-driven pixel classification. However, the fixed weights of CNNs make it difficult for them to dynamically adapt to the large intra-class variance in remote sensing images. Although Transformers can model global context, their indiscriminate attention mechanism is particularly sensitive to highly similar inter-class features and, lacking explicit intra-class constraints, is prone to causing class confusion \cite{78}.
	
	These inherent limitations have prompted researchers to recognize that class constraints must be shifted from the end-stage classification task to the upfront feature extraction process. Driven by this idea, class-guided methods have emerged, with their core strategy being the modeling of class semantics to guide feature extraction, thereby compelling the model to focus on features relevant to class semantics \cite{21, 22, 23, 75,93}.
	
	Although these methods have achieved significant success, they still fail to fully consider the unique characteristics of remote sensing imagery. Firstly, they rely on coarse-grained class prototypes in the initial stage to guide the extraction of global contextual features. These initial coarse class prototypes can hardly represent class features comprehensively and  introduce misleading characteristics. Secondly, they typically depend on class weights and self-attention mechanisms to establish the association between class information and features. Strategies based on class weights lose spatial details during information compression and are unable to establish contextual correlations. Moreover, the self-attention mechanisms subsequently used to establish these contextual links perform poorly in both precision and efficiency when distinguishing between similar features, owing to their inherent smoothing effect and quadratic computational complexity \cite{40, 41, 42,94,95}.

	The recently proposed Mamba model \cite{25} efficiently captures long-range dependencies with linear complexity, offering an alternative to the computationally expensive self-attention mechanism. Researchers have found that by designing the order in which features enter the SSM, the model can be guided to focus on different types of information\cite{26}.
	The exploration in some recent Mamba variants has primarily focused on enhancing spatial or modal feature interaction by varying the scan order \cite{28, 37, 77, 69,96,97}.
	However, these methods generally lack a crucial mechanism for translating Mamba's sequential modeling capabilities into a refined, class-information-driven feature selection strategy. More importantly, existing class-guided methods prioritize the consistency of class semantics while neglecting structural information.
	
	In this paper, we introduce the Prototype-Driven Structure-Synergized Network (PDSSNet), which consists of a feature encoder, an Adaptive Prototype Extraction Module (APEM), a Semantic-Structure Coordination Module (SSCM), and a Channel Similarity Adjustment Module (CSAM). Conventional class modeling approaches typically rely on the segmentation head to produce coarse-grained results, leading to class prototypes that struggle to comprehensively represent the high intra-class variance found in remote sensing images.
	The proposed APEM overcomes this limitation by encoding ground truth to create a binary mask for each class. This mask facilitates the aggregation of same-class features across multiple channels using its “1” values, while the “0” values simultaneously suppress the influence of features from other classes. This mechanism allows for the generation of comprehensive class prototypes.
	
	Upon obtaining precise class prototypes, how to leverage them as effective prior guidance becomes another critical task. We think that for a single ground object class, its class semantics are unified and invariant, whereas its spatial structure is concrete and variable. However, the design of existing class-guided methods generally tends to reinforce semantic consistency while relatively neglecting the constraint on target structural integrity. This lack of constraint on structural integrity inevitably compromises the completeness of the segmentation.
	To this end, we propose the SSCM to enhance the integrity of class information. This module simulates the hierarchical cognitive strategy of humans: it first embeds class-level semantic information to enable the model to comprehend global semantics, and then progressively refines class details based on extracted structural information, allowing the model to dynamically adjust class representations according to structural differences.
	To effectively address the challenge of high inter-class similarity, we have designed the CSAM. This module, based on a core  SimStep mechanism, enables the model to bypass local similarities by controlling its step size when encountering similar features, focusing instead on discriminative characteristics. Experimental results validate the effectiveness of the proposed method.In summary, the main contributions of this study are as follows.

	\begin{enumerate}
		\item To overcome the high intra-class variance in remote sensing images, we propose a new class modeling method, the APEM. It aggregates diverse features under a single class by leveraging the ground truth to construct a comprehensive class prototype.
		\item We design the SSCM, the core of which is to drive the dynamic alignment of high-level semantic understanding with spatial structure layout. It utilizes the continuity of structural information to constrain the scope of semantic prediction, ensuring that the semantic guidance can adaptively conform to the complete contours of a class in its various morphologies.
		\item We propose the PDSSNet, a network that refines the semantic and structural representations of classes by integrating APEM and SSCM to ensure their morphological integrity. Building on this, we have designed the CSAM to enhance the final discriminative capability for highly similar classes, thereby comprehensively addressing the problem of incomplete class segmentation.	
	\end{enumerate}

	\section{RELATED WORK}
	\subsection{Class-guided Semantic Segmentation of Remote Sensing Images}
	To address the challenges of traditional methods in learning unified and discriminative class representations\cite{16},\cite{58},\cite{60}, \cite{61},\cite{62} researchers have proposed class-guided methods. The core idea is to provide more explicit guidance for the feature extraction and optimization processes by explicitly modeling class semantic information within the network, thereby enhancing segmentation accuracy.
	
	Most existing class-guided methods adhered to a common paradigm: they utilized coarse-grained segmentation maps from the network's intermediate layers as dynamic class priors. For example, CGGLNet \cite{22} generated coarse segmentation results at different network stages to guide the extraction of global contextual information, while CGGCNet \cite{78} took this a step further. It used an adjacency matrix, calculated from the coarse segmentation matrix, to enforce tight connections among same-class pixels and explicitly model contextual relationships.
	Other works optimized class representations from different perspectives. LoG-CAN \cite{33} integrated class context through cascaded refinement and feature fusion, and its successor, LoG-CAN++ \cite{88}, introduced affine transformations to extract local class centers. Furthermore, some methods performed modeling by mining deeper intra- and inter-class relationships. For instance, CGCWNet \cite{79} modeled high-order correlations between classes and channels using second-order feature statistics, and CAGNet \cite{21} generated class attention weights by constructing a class matrix and embedding positional information.
	Despite the significant progress achieved by these methods, their guiding prototypes were derived from the model's own coarse-grained predictions. When faced with high intra-class variance, these coarse prototypes had limited representational capacity and inevitably introduced misleading features\cite{80,89}.
	
	In contrast, our APEM completely abandons the traditional path of relying on model predictions. APEM utilizes ground truth to directly aggregate all diverse representations of a single class from the feature maps, thereby constructing a comprehensive semantic prototype. This approach precludes the introduction of misleading information at its source. To further enhance the adaptability of the class prototypes to complex scenes, we leverage the enhanced features from the SSCM's output to dynamically update these prototypes.

	\subsection{State-Space Models}
	In recent years, the rapid development of State Space Models (SSMs) in the field of deep learning has significantly enhanced the capability for long-range dependency modeling and sequential data processing. The introduction of HiPPO initialization has facilitated the integration of SSMs with deep learning, strengthening their ability to capture long-range dependencies\cite{71}. The LSSL model first addressed the challenge of long-term dependence in SSMs\cite{72}, but it still has limitations in computational and memory efficiency. Subsequently, the S4 model\cite{73} reduced computational overhead through a novel parameterization strategy, making SSMs more practical. On this basis, SSMs have derived various variants, such as those using complex diagonal structures to enhance temporal modeling\cite{85} or supporting multi-input multi-output configurations to improve flexibility\cite{50}.

	\begin{figure*}[!t]
		\centering
		\includegraphics[width=7in]{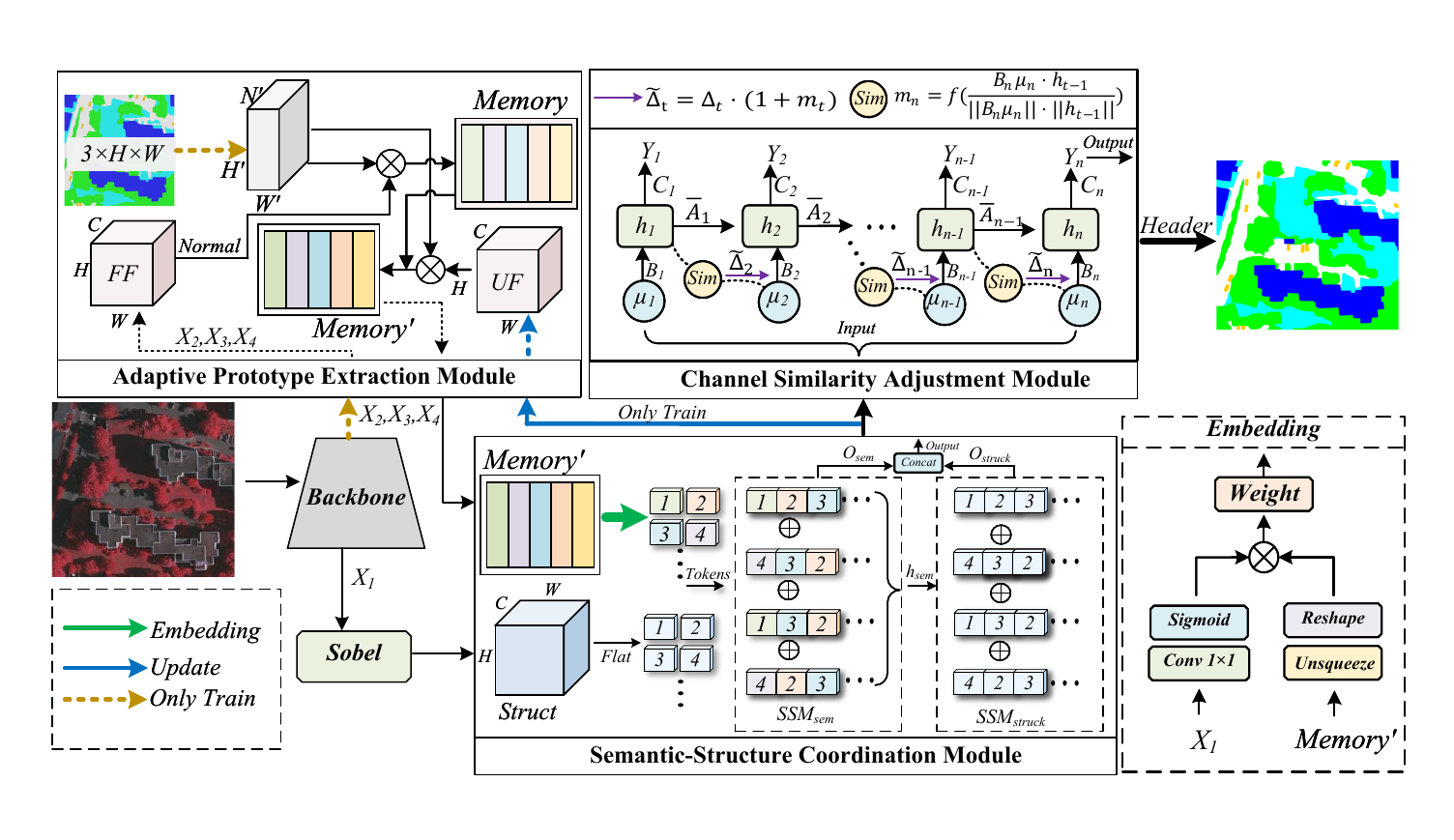}
		\vspace{-30pt}
		\caption{The architecture of PDSSNet. First, the input image undergoes downsampling, and features from the last three stages are extracted to mix and initialize the APEM. Next, the prototype features output by the APEM and the structural information obtained via the \(Sobel\) operator are enhanced through feature interaction in the SSCM. The enhanced features are then split into two branches: one branch reversely updates the prototype representation, while the other branch is processed by the CSAM, after which the segmentation head outputs the final segmentation result.}
		\label{fig2}
	\end{figure*}
	
	\begin{figure}[!t]
		\includegraphics[width=3.5in]{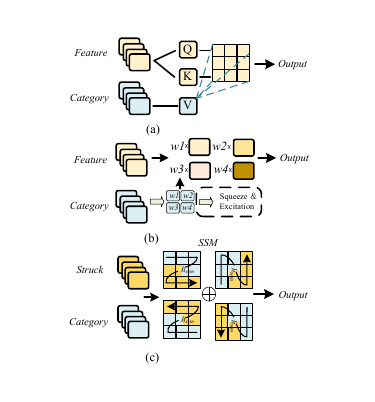}
		\vspace{-45pt}
		\caption{Three different class-guiding methods are compared. (a) Traditional self-attention method. (b) Generating class weights. (c) Our proposed F-SS2D hierarchical guidance method.
		}
		\label{fig3}
	\end{figure}

	In the field of remote sensing, SSMs have attracted significant attention due to their long-range modeling capabilities and linear complexity.
	A recent research direction has been to extend the powerful capabilities of one-dimensional sequence processing for application to remote sensing images. RS-Mamba \cite{28} was the first to apply SSMs to remote sensing dense prediction tasks, validating their feasibility. RSCaMa \cite{37} designed the TT-SSM module by leveraging the potential correlation between Mamba's temporal scanning characteristics and the temporality of the RSICC task, which effectively enhances feature interaction across different timestamps. FusionMamba \cite{69} fused information from different modalities by extending its unidirectional input to be bidirectional. Meanwhile, CMS2I-Mamba \cite{77} employed a multi-path selective scanning mechanism to promote the fusion of spectral and spatial features. Furthermore, models such as Samba \cite{30} and PyMamba \cite{27} demonstrated the impressive performance of the original Mamba block.
	
	However, most of these works are limited to spatial or modal feature interaction, failing to effectively combine Mamba's advantages in sequential modeling with the guiding role of class semantics. In remote sensing image segmentation, class semantics are core to distinguishing different ground objects, while structural information helps in generating consistent segmentation results. Existing methods often overlook the synergistic effect between these two aspects.
	To this end, we propose the SSCM module, which adopts a hierarchical cognitive strategy to achieve a synergy between class semantics and structural information, thereby enhancing the model's ability to segment complete class instances.

	\section{METHODOLOGY}
	In this section, we first provide an overview of the overall architecture of PDSSNet. Subsequently, we will delve into the key components of PDSSNet, including APEM, SSCM, and CSAM, elaborating on their respective roles in addressing the problem of incomplete category segmentation.
	
	\subsection{PDSSNet Architecture}
	
	The architecture of PDSSNet is shown in Fig. \ref{fig2}. We adopt a pre-trained ConvNext\cite{39} as the backbone network. ConvNext consists of four stages, with each stage performing downsampling.
	For an input image \(x \in \mathbb{R}^{3 \times h \times w}\), we first perform four downsampling operations to obtain spatial features at four different scales: \(x_1 \in \mathbb{R}^{C \times H \times W}\), \(x_2 \in \mathbb{R}^{2C \times (H/2) \times (W/2)}\), \(x_3 \in \mathbb{R}^{4C \times (H/4) \times (W/4)}\), and \(x_4 \in \mathbb{R}^{8C \times (H/8) \times (W/8)}\).
	Next, interpolation and convolution operations are applied to fuse \(x_2\), \(x_3\), and \(x_4\):
	\begin{equation}
		FF = \text{Cat} \left( \sigma_{1 \times 1} \left( \xi(x_2) \right), \sigma_{1 \times 1} \left( \xi(x_3) \right), \sigma_{1 \times 1} \left( \xi(x_4) \right) \right),
		\label{eq:1}
	\end{equation}
	where $\xi(\cdot)$ denotes the interpolation operation, $\sigma_{k\times k}(\cdot)$ signifies the convolution operation with a kernel of size \( k\times k \). The fused features \(FF\in\mathbb{R}^{3C\times H\times W}\), \(FF\) initializes the storage matrix \(Memory \in \mathbb{R}^{N \times C}\) (abbreviation:\(M\)), where \(N\) is the number of classes and \(C\) is the channel dimension of the encoder features Specifically in Section \hyperref[apem-module]{B.2)}.
	Meanwhile, structural information is generated for \(x_1\) through \(Sobel\) operations \( Struct\in R^{C\times H\times W}\).
	
	Next, before feeding the obtained category semantic and structural information into SSCM, we perform simple spatial position adjustment on the category semantic information to facilitate SSCM's capture of spatial dependencies between categories.The relevant formula is as follows:
	\begin{equation}\label{eq:reshape_operation}
		Em = \text{Reshape}(\text{Unsequeeze}(M)),
	\end{equation}
	\begin{equation}\label{eq:sigmoid_operation}
		\mathcal{W} = \text{Sigmoid}(\text{Conv}_{1 \times 1}(\textit{\(x_1\)})),
	\end{equation}
	\begin{equation}\label{eq:position_weighting}
		\text{\(Pos\)}[n \cdot C : (n + 1) \cdot C, i, j] = \mathcal{W}[n, i, j] \cdot Em [n],
	\end{equation}
	where \(E_m \in \mathbb{R}^{N \cdot C \times H \times W}\), \(\mathcal{W}[n, i, j]\) represents the confidence weight of class \(n\) at position \((i, j)\), and \(Pos\) represents the regionalized features of \(N\) classes at position \((i, j)\).

	Thereafter, category semantic information integrated with spatial positions \(Pos\) and structural information \(Struck\) is processed by SSCM’s hierarchical cognitive strategy to generate enhanced features (specifically described in Section \hyperref[sscm-module]{B.3)}). These enhanced features are divided into two branches: one branch reversely updates the memory matrix \(M\) to enrich the category prototype representation, and the other branch is fed into the CSAM. Through the designed SimStep mechanism, CSAM increases the update step size of hidden states when encountering similar features, enabling the model to focus on discriminative features. Finally, the output of the CSAM is fed back to the segmentation head to generate high-precision segmentation results.
	
	By integrating the Adaptive Prototype Extraction Module and Semantic-Structure Coordination Module, PDSSNet dynamically adapts to complex ground objects in remote sensing images, significantly enhancing the model’s category parsing capability. APEM provides accurate category semantic information, while SSCM simulates the human hierarchical cognitive strategy of “semantic understanding first, then structural refinement.” Additionally, PDSSNet employs CSAM to further improve the model’s ability to discriminate confusing features. Through the synergistic effect of these modules, PDSSNet has effectively enhanced the category parsing ability in complex scenes and improved the completeness of category segmentation results.

	\subsection{Decoder}
	In this section, we will introduce the implementation of APEM, SSCM, and CSAM. Before that, we will first introduce the preliminary knowledge of SSM.

	\subsection*{\itshape 1) State - Space Model}The SSM is known as a linear time-invariant system, which maps the input sequence \(\mu(t) \in \mathbb{R}\) to the response sequence \(y(t) \in \mathbb{R}\) through \(h(t) \in \mathbb{C}^N\). These mappings are typically represented by
	\begin{equation}\label{eq:state_space}
		h'(t) = Ah(t) + B\mu(t),
	\end{equation}
	\begin{equation}\label{eq:y state_space}
		y(t) = Ch(t) + D\mu(t).
	\end{equation}
	
	Equation (\ref{eq:state_space}) is the state transition equation, and Equation (\ref{eq:y state_space}) is the output equation. \(A \in \mathbb{C}^{N \times N}\) is the state transition matrix, \(B \in \mathbb{C}^N\) is the input control matrix, \(C \in \mathbb{C}^N\) is the output mapping matrix, and \(D \in \mathbb{C}^1\) is the skip connection matrix. Given the time step \(\Delta \in \mathbb{R}\), the continuous-time model is discretized using the Zero-Order Hold (ZOH) method, resulting in:
	\begin{equation}\label{eq:ht}
		h_t = \bar{A}(h)_{t - 1} + \bar{B} \mu_t,
	\end{equation}
	\begin{equation}\label{eq:yt}
		y_t = C h_t + D \mu_t,
	\end{equation}
	\begin{equation}\label{eq:A_bar}
		\bar{A} = e^{\Delta A},
	\end{equation}
	\begin{equation}\label{eq:B_bar}
		\bar{B} = (\Delta A)^{-1} (e^{\Delta A} - I) \Delta B.
	\end{equation}
	
	For Equation (\ref{eq:B_bar}), it can be obtained by using the first-order Taylor expansion:
	\begin{equation}\label{11}
		\bar{B} \approx (\Delta A)^{-1}(\Delta A)\Delta B = \Delta B.
	\end{equation}
	
	By combining the above Equations (\ref{eq:ht}), (\ref{eq:yt}), (\ref{eq:A_bar}) and Equations (\ref{11}), the output \(y\) can be derived and expressed as \(y=\left[ y_1, y_2, \ldots, y_t, \ldots, y_L \right]\).

		%
		%
		%

		\subsection*{\itshape 2) Adaptive Prototype Extraction Module}
		\hypertarget{apem-module}{\label{apem-module}}
		In remote sensing image segmentation tasks, class information is crucial for feature extraction, especially in scenes with complex ground objects. Existing methods, such as CGGLNet \cite{22}, CGGCNet \cite{78}, and CG-Swin \cite{23}, guide feature extraction by generating coarse-grained class prototypes in their intermediate layers. These coarse-grained prototypes struggle to effectively handle the large intra-class variance. In contrast, our approach utilizes one-hot encoded ground truth, enabling the generation of comprehensive class prototypes without introducing misleading features.
		The following details the initialization and update processes of the APEM.
		
		\noindent\textbf{Initialization.} To generate initial representative class prototypes, we perform pixel-level masking on the feature map \(FF\) based on the ground truth segmentation map and calculate the average feature vector for each category to form a preliminary class prototype representation. Specifically, for the input \(FF\) \(\in \mathbb{R}^{3C \times H \times W}\), we obtain \(\mathcal{Q} \in \mathbb{R}^{C \times H' \times W'}\) through \(\ell_2\) regularization and \(1 \times 1\) convolution. For the ground truth \(\text{\(Y\)} \in \mathbb{R}^{3 \times h \times w}\), we perform one-hot encoding and bilinear interpolation to generate \(\mathcal{L} \in \mathbb{R}^{N \times H' \times W'}\) with the same scale as \(\mathcal{Q}\). To initialize the \(k\)-th entry \(M[k]\) in the storage matrix, we perform average pooling on the masked region by referring to the segmentation mask of the \(k\)-th class, as shown below:
		\begin{equation}
			M[k] = \frac{\alpha_k \cdot \mathcal{L}[k] \mathcal{Q}^{\mathrm{T}}}{\sum \alpha_k \cdot C_k}, \quad
			\alpha_k = \frac{1}{\log(C_k + 1)},
		\end{equation}
		where \(C_k\) is the number of pixels belonging to the k-th class in the ground truth, \(M[k] \in \mathbb{R}^{N \times C}\) is the matrix storing class prototypes, \(\mathcal{L}\) is the one-hot encoded ground truth with size \(N \times H' W'\), and \(\mathcal{Q}\) is reshaped into \(C \times H' W'\). Considering the extremely imbalanced class distribution in remote sensing images as shown in Fig. \ref{fig5}, we introduce a class weight \(\alpha_k\) to ensure the contribution of rare classes.
		
		\noindent\textbf{Update.} 
		To evolve the class prototypes from a static benchmark into a dynamic representation adaptable to complex scenes, we utilize the enhanced features \(UF \in \mathbb{R}^{C\times H\times W}\) output by the SSCM to dynamically update them.
		\begin{equation}
			\tilde{\mathcal{A}}[k] = \mathcal{L}[k]UF^{\prime\mathrm{T}},
		\end{equation}
		\begin{equation}
			M^{\prime}[k] = \bm{\beta} \cdot M[k] + (1 - \bm{\beta}) \cdot \tilde{\mathcal{A}}[k],
		\end{equation}
		where \(\widetilde{\mathcal{A}}[k] \in \mathbb{R}^{N \times C}\) is the masked feature map, \({UF}'\) is obtained by performing \(\ell_2\) regularization on \(UF\) and reshaped into \(C \times H' W'\), \(M'[k]\) is the updated matrix, and the parameter \(\bm{\beta}\) is set to 0.7.

		\begin{figure*}[!t]
			\centering
			\includegraphics[width=7in]{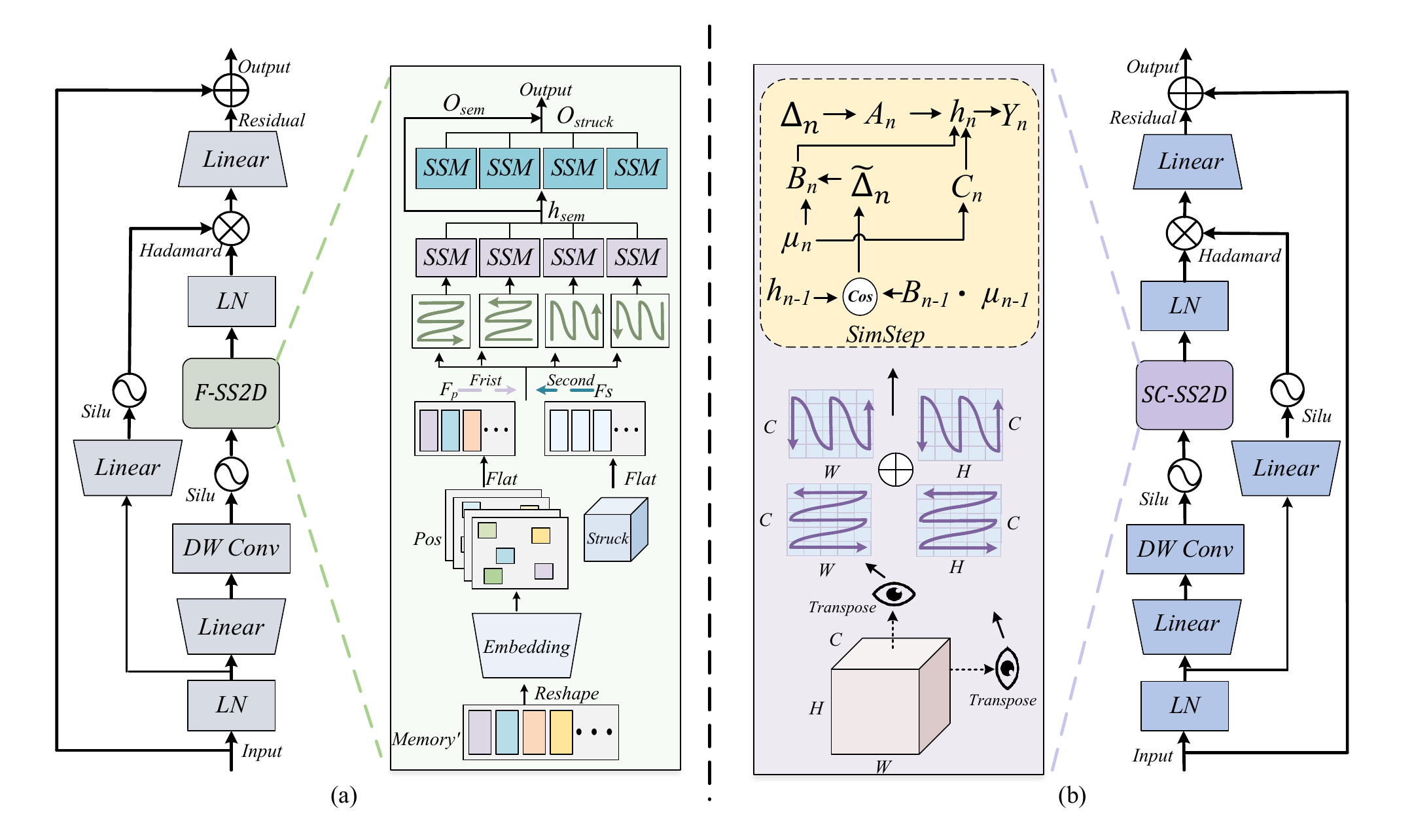}
			\vspace{-15pt}
			\caption{The structures of SSCM and CSAM.
				(a) The structure of SSCM takes F-SS2D as the core component.
				(b) The structure of CSAM takes SC-SS2D as the core component.}
			\label{fig4}
			
		\end{figure*}

		\subsection*{\itshape 3) Semantic-Structure Coordination Module }
		\hypertarget{sscm-module}{\label{sscm-module}}

		As illustrated in Fig. \ref{fig3}, most previous category-guided methods rely on self-attention frameworks or generate category weights by compressing category information. The category weight generation approach depends on global spatial statistics, ignoring local details and leading to spatial information loss. Methods based on self-attention frameworks use global \(QKV\) computation and Softmax to generate weights, but Softmax tends to homogenize weights, reducing discriminability. In high-resolution images, this results in “blurred” feature representations and feature guidance errors \cite{40,41,42}. Additionally, these methods often overlook class structural information, and this lack of structural constraint leads to the fragmentation of target morphologies.
		To address this, we designed the SSCM to enable dynamic interaction between semantic and structural information. Inspired by human cognitive laws—where humans first abstract category concepts through long-term memory when perceiving objects, then refine cognition based on observed details, integrate global semantic relationships, and finally iteratively optimize category understanding through feedback \cite{43,44} SSCM improves the expression of class information by simulating the dynamic cognitive process of humans in complex scenes.
		
		The VSS block in Vmamba can effectively establish dependency relationships in the spatial domain, and we choose to incorporate VSS. As shown in the Fig. \ref{fig4}.
		The core component of SSCM, F-SS2D, consists of three main parts: the hierarchical cognitive spatial scanning algorithm (HS), SSM establishment, and merging operations.
		For the input spatially modulated semantic tensor \(Pos\) and structural information tensor \(Struct\), we first serialize them into one-dimensional sequences, denoted as \(F_p\) and \(F_s\), respectively. 
		The semantic sequence \(F_p\) is first processed by the State Space Model \(SSM_\text{sem}\) to establish global contextual dependencies and capture its final hidden state, \(h_\text{sem}\).
		This captured state \(h_\text{sem}\) is then used as the initial state for the State Space Model, \(SSM_\text{struct}\). This ensures that the processing of the structural sequence \(F_s\) is conditioned on the prior knowledge established during the semantic scan.
		The entire hierarchical process can be expressed by the following set of equations:
		
		\begin{equation}
			(O_{\text{sem}, k}, h_{\text{sem}, k}) = \text{SSM}_{\text{sem}}(F_{p, k}),
		\end{equation}
		\begin{equation}
			O_{\text{struct}, k} = \text{SSM}_{\text{struct}}\bigl(F_{s, k}, \text{initial\_state} = h_{\text{sem}, k}\bigr),
		\end{equation}
		\begin{equation}
			Output = \text{Concat}(O_{\text{sem}}, O_{\text{struct}}),
		\end{equation}
		where \(k \in \{\text{h}, \text{v}, \text{hr}, \text{vr}\}\) represents one of the four scan directions (horizontal, vertical, and their respective reverse directions). \(O\) represents the output features.
		\subsection*{\itshape 4) Channel Similarity Adjustment Module}
		The similarity between category channels is the main cause of category confusion in remote sensing image segmentation \cite{74}.
		To strengthen the model's capacity to differentiate between similar channels,  inspired by RWKV\cite{48} and Hyena\cite{49}, a SimStep mechanism based on the SSM is proposed to control the state update of similar channels by dynamically driving the step size through similarity. Specifically, this mechanism dynamically adjusts the step size by measuring the similarity between the hidden state and the current input: increasing the step size allows the model to “skip” or “compress” short-term similar features, tend to update the state quickly, and weaken the cumulative effect of short-term similar inputs, so as to pay more attention to significant changes or long-term dependencies in the sequence. It is worth noting that although Mamba has an implicit filtering mechanism, it focuses more on general sequence modeling and long-term dependencies, without explicit processing of similarity\cite{25}.
		Cosine similarity is employed to quantify the similarity between feature vectors\cite{68}.
		
		The structure of CSAM is similar to that of SSCM, but it adopts SC-SS2D instead of the F-SS2D algorithm in the channel dimension. SC-SS2D consists of the spatial scanning algorithm CS\cite{32} and the SimStep mechanism designed by us. As shown in the Fig. \ref{fig4}. In the channel modeling approach of the SimStep mechanism, we only process Equation (\ref{11})  to dynamically adjust the update of historical states according to different inputs. Specifically:
		\begin{equation}
			\widetilde{\Delta}_t = \Delta_t \cdot (1 + m_t),
		\end{equation}
		\begin{equation}
			m_t = \text{Sigmoid}(\text{ReLU}(W_s \cdot s_t + b_s)),
		\end{equation}
		\begin{equation}
			s_n = \frac{B_n \mu_n \cdot h_{t - 1}}{\| B_n \mu_n \| \cdot \| h_{t - 1} \|},
		\end{equation}
		where \(W_s\) and \(b_s\) are the weights and biases of the linear layer, respectively. In the selection of nonlinear functions for the SimStep mechanism, we prioritize the ReLU function, whose ability to distinguish between positive and negative features enables the identification of similar features requiring larger step sizes. In terms of temporal complexity, \(s_t\) depends on the previous hidden state \(h_{t-1}\). In Mamba's selective scanning, all values of \(h_{t-1}\) are precomputed and cached via parallelization techniques. Therefore, the computation of \(s_t\) can be integrated into the scanning process without waiting for results from the previous time step, enabling full parallelization\cite{50,52}. To ensure that our category storage matrix contains more detailed category information, the features after the CSAM do not update the category matrix.
		
		\subsection{Loss Function}
		For the PDSSNet proposed by us, the entire network is trained using a loss function \(\mathcal{L}\), which consists of \(\mathcal{L}_{ce}\) and \(\mathcal{L}_{dice}\). Specifically:
		
		\begin{equation}
			\mathcal{L}_{ce} = -\frac{1}{N} \sum_{n = 1}^{N} \sum_{k = 1}^{K} y^{(n)}_k \log \hat{y}^{(n)}_k,
		\end{equation}
		\begin{equation}
			\mathcal{L}_{dice} = 1 - \frac{2}{N} \sum_{n = 1}^{N} \sum_{k = 1}^{K} \frac{\hat{y}^{(n)}_k y^{(n)}_k}{\hat{y}^{(n)}_k + y^{(n)}_k},
		\end{equation}
		\begin{equation}
			\mathcal{L} = \mathcal{L}_{ce} + \mathcal{L}_{dice},
		\end{equation}
		where \(N\) denotes the number of training samples and \(K\) denotes the number of classes. \(y_k^{(n)}\) represents the one-hot encoding of the ground truth that sample \(n \in [1, \ldots, N]\) belongs to class \(k\), and \(\hat{y}_k^{(n)}\) denotes the probability that sample n belongs to class \(k\).

		\section{EXPERIMENTS}
		
		\subsection{Experimental Settings}
		
		\subsubsection{Datasets}
		For our evaluation, we used the Vaihingen, Potsdam, and LoveDA datasets, which are standard benchmarks for remote sensing semantic segmentation. These widely-used datasets were selected to ensure the comparability and generalizability of our results, as they cover diverse land categories, such as buildings and road systems, under complex environmental conditions like varied seasons and lighting. Their prevalence in academia facilitates a direct comparison of our work with previous studies.

		\paragraph{Vaihingen dataset}
		The dataset comprises 33 high-resolution TOP image tiles, each with an average dimension of 2494$\times$2064 pixels, featuring real orthophoto (TOP), digital surface model (DSM), and normalized DSM (NDSM) data. It includes five foreground land cover categories: impervious surfaces, buildings, low vegetation, trees, and cars, along with a background class: clutter. During the experiment, we carefully chose training samples based on the specific training IDs (ID 1, 3, 5, 7, 11, 13, 15, 17, 21, 23, 26, 28, 30, 32, 34, and 37) provided by the ISPRS competition, using the remaining 17 images for testing. To facilitate processing, these image tiles were cropped into smaller 1024$\times$1024 pixel blocks.
		
		\begin{figure}[!t]
			\centering
			\includegraphics[width=3.5in]{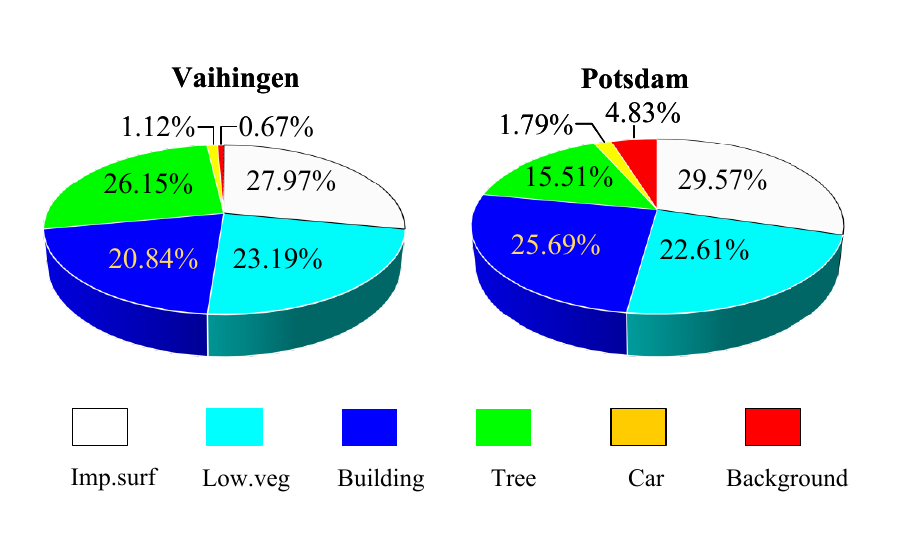}
			\caption{Datasets’ label proportion chart.
			}
			\label{fig5}
		\end{figure}
		
		\paragraph{Potsdam dataset}
		The Potsdam dataset offers a more extensive and challenging evaluation scenario, containing 38 high-resolution tiles from Potsdam, Germany. Each 6000$\times$6000 pixel tile features a fine-grained ground sampling distance (GSD) of 5 cm and provides data across four multispectral bands (Red, Green, Blue, and Near-Infrared), accompanied by a corresponding Digital Surface Model (DSM). According to the official ISPRS protocol, the dataset is divided into a training set composed of 24 specific tiles (IDs: 2\_10, 2\_11, 2\_12, 3\_10, 3\_11, 3\_12, 4\_10, 4\_11, 4\_12, 5\_10, 5\_11, 5\_12, 6\_7, 6\_8, 6\_9, 6\_10, 6\_11, 6\_12, 7\_7, 7\_8, 7\_9, 7\_11, and 7\_12) are utilized for training, leaving the other 15 tiles for the test set. For processing, these large tiles are partitioned into 1024$\times$1024 patches. Following common practice, the 'clutter/background' class is excluded during quantitative evaluation.
		
		\paragraph{LoveDA dataset}
		The dataset is constructed from Google Earth imagery, spanning three Chinese cities: Nanjing, Changzhou, and Wuhan. It comprises 5,987 high-resolution optical remote sensing images at 0.3 meter ground resolution. Each 1024$\times$1024 pixel image includes seven land cover categories: buildings, roads, water bodies, wasteland, forests, agricultural areas, and background. The dataset contains three spectral bands—red, green, and blue\cite{54}.
		Unlike Vaihingen and Potsdam dataset processing, LoveDA divides training, validation, and test sets by geographical location to amplify regional diversity. Specifically, the dataset is split into 2,522 training images, 1,669 validation images, and 1,796 official test images. It encompasses both urban and rural landscapes. However, the dataset poses significant research challenges due to multi-scale objects, complex backgrounds, and imbalanced class distributions.
		
		\subsubsection{Implementation Details}
		All experiments were conducted on a single NVIDIA GeForce RTX 2080Ti (11 GB) GPU, running on an Ubuntu 18.04 system with the PyTorch 1.11 framework. For model optimization, we employed the AdamW optimizer, setting the initial learning rate to 6e-4 and decaying it throughout training via a cosine scheduling strategy.
		For the Vaihingen and Potsdam datasets, all images were pre-processed by partitioning them into 1024$\times$1024 patches. During the training phase, a suite of data augmentation techniques was applied to enhance model generalization, including random horizontal/vertical flipping, random rotation, and random scaling with factors of {0.5, 0.75, 1.0, 1.25, 1.5}. To ensure a robust evaluation, the testing phase incorporated a multi-scale assessment strategy combined with random flip augmentation.

		\subsubsection{Evaluation Metrics}
		To quantitatively evaluate the effectiveness of the proposed model, we adopt three metrics: Overall Accuracy (OA), Mean Intersection over Union (mIoU), and F1 score (F1). The relevant formulas are as follows:
		\begin{equation}
			\text{OA} = \frac{\sum_{n = 1}^{N} \text{TP}_n}{\sum_{n = 1}^{N} \text{TP}_n + \text{FP}_n + \text{TN}_n + \text{FN}_n},
		\end{equation}
		\begin{equation}
			\text{mIoU} = \frac{1}{N} \frac{\sum_{n = 1}^{N} \text{TP}_n}{\sum_{n = 1}^{N} \text{TP}_n + \text{FP}_n + \text{FN}_n},
		\end{equation}
		\begin{equation}
			\text{precision}_n = \frac{\text{TP}_n}{\text{TP}_n + \text{FP}_n}, \quad
			\text{recall}_n = \frac{\text{TP}_n}{\text{TP}_n + \text{FN}_n},
		\end{equation}
		\begin{equation}
			\text{F1} = 2 \times \frac{\text{precision}_n \times \text{recall}_n}{\text{precision}_n + \text{recall}_n},
		\end{equation}
		where \(\mathrm{TP}_n\), \(\mathrm{FP}_n\), \(\mathrm{TN}_n\), and \(\mathrm{FN}_n\) represent the true positive count, false positive count, true negative count, and false negative count, respectively, all of which are associated with the specific object of category \(n\) (indexed by \(n\)).

		\subsection{Ablation Experiment}
		To assess the effectiveness of PDSSNet's individual components, we performed experimental evaluations on the Vaihingen, Potsdam, and LoveDA datasets. The experimental results are averaged over multiple trials and analyzed primarily through two performance metrics: mIoU  and F1

		\subsubsection{Components of PDSSNet}
\begin{figure}[!t]
			\centering
			\includegraphics[width=3.5in]{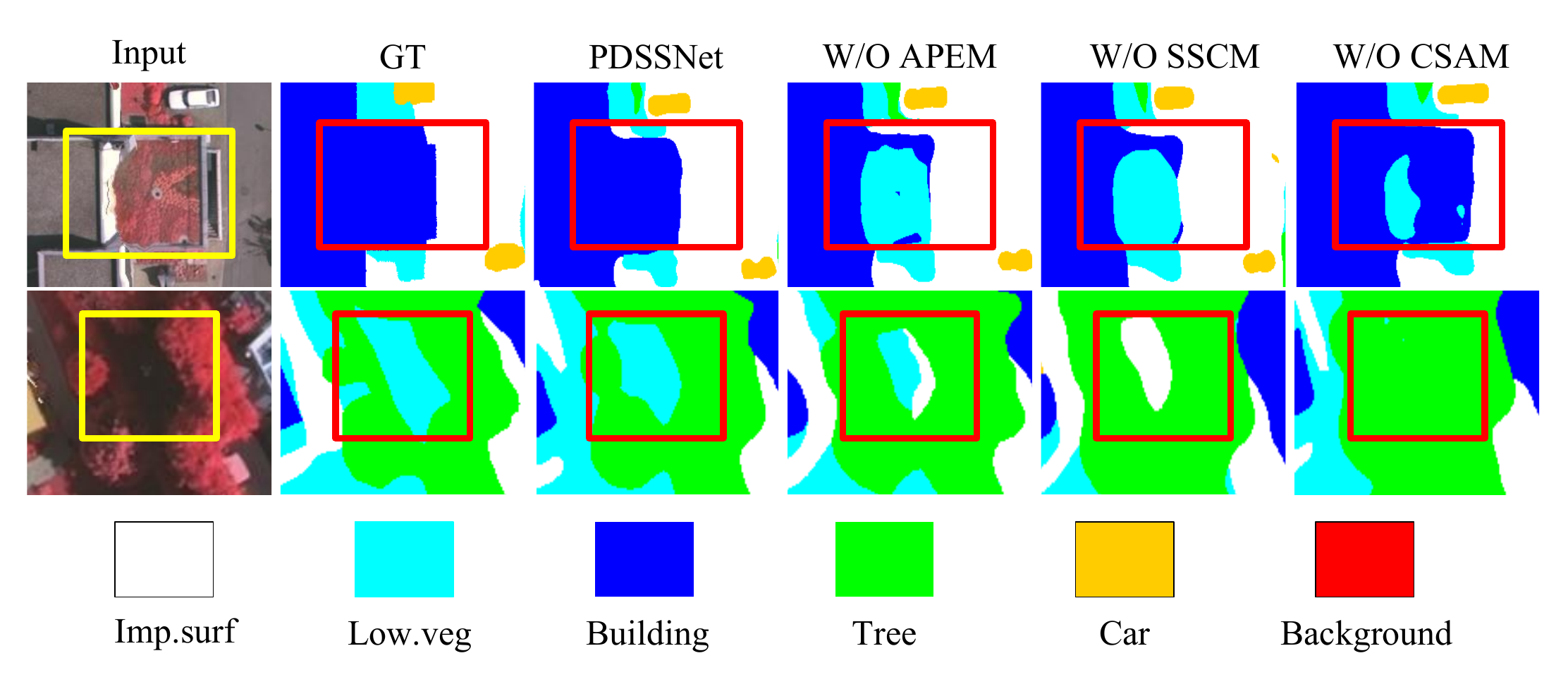}
			\caption{Partial image of PDSSNet with one module removed on the Vaihingen dataset.
			}
			\label{fig6}
		\end{figure}

\begin{figure}[!t]
			\centering
			\includegraphics[width=3.5in]{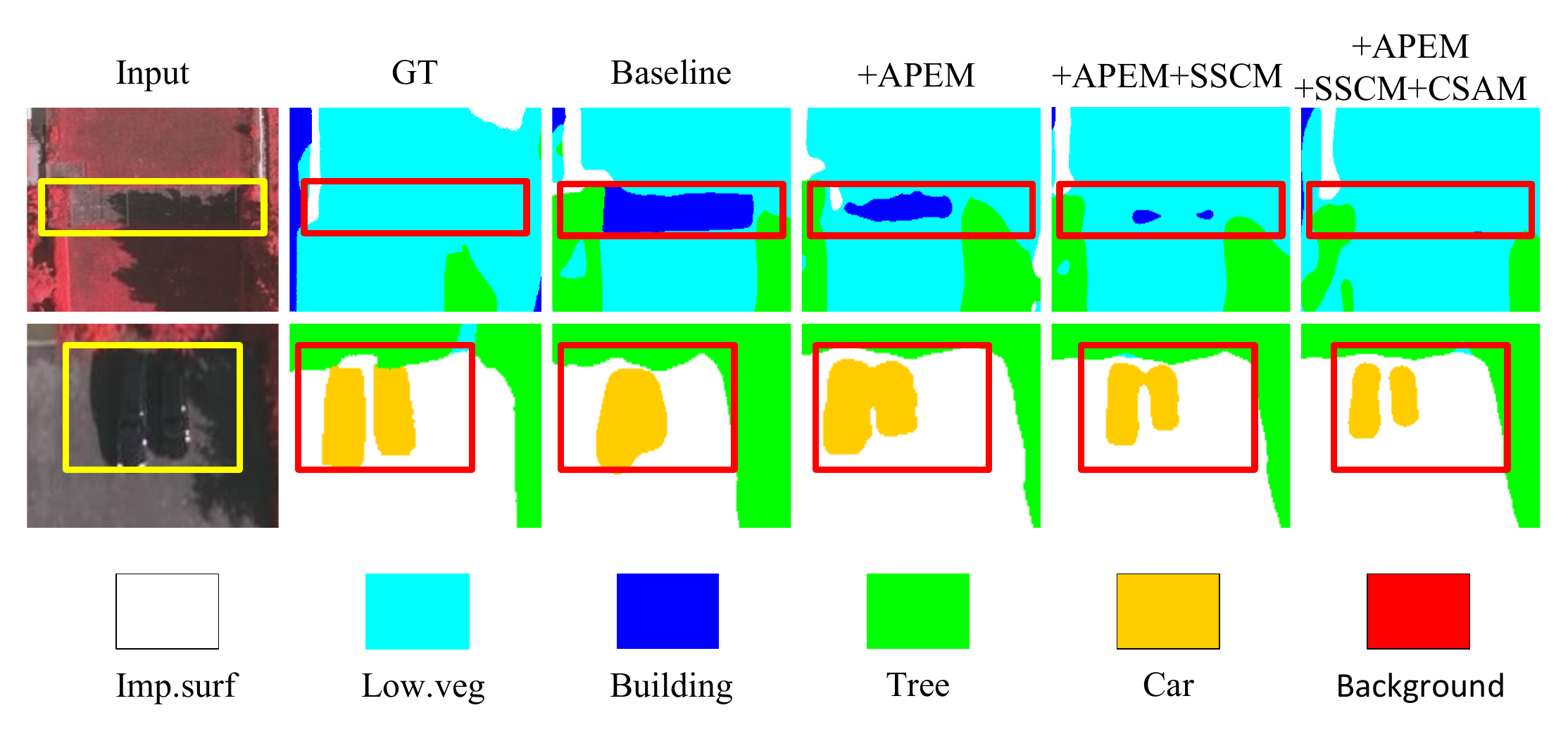}
			\caption{Baseline adds APEM, SSCM, and CSAM partial enlargement pictures accordingly.
			}
			\label{fig7}
		\end{figure}

		To investigate the role of each component in PDSSNet, we conducted a series of experiments. Table \ref{tab1} displays the experimental results after module-wise ablation, where APEM represents the Adaptive Prototype Extraction Module, SSCM represents the Semantic-Structure Coordination Module, and CSAM represents the Channel Similarity Adjustment Module, with (w/o) indicating the removal of a component and (+) indicating its inclusion in the experiments. Fig. \ref{fig6} presents the segmentation performance after deleting a single module of PDSSNet, demonstrating that removing any component negatively impacts network performance.
		To deeply evaluate the effectiveness of APEM, SSCM, and CSAM, we performed special experiments on the Vaihingen dataset, with the results presented in Table \ref{tab7}. These experiments involved using a baseline model (consisting of two convolutional layers connected directly after the ConvNext feature extraction network with a segmentation head) and then individually inserting APEM, SSCM, and CSAM before the segmentation head.
		Fig. \ref{fig7} shows the segmentation results after sequentially adding each component to the baseline model, and comprehensive experimental data indicate that each newly proposed component can significantly improve model performance. Table \ref{tab3} gives the experimental results of replacing different backbones.

		\begin{figure}[!t]
			\centering
			\includegraphics[width=3.5in]{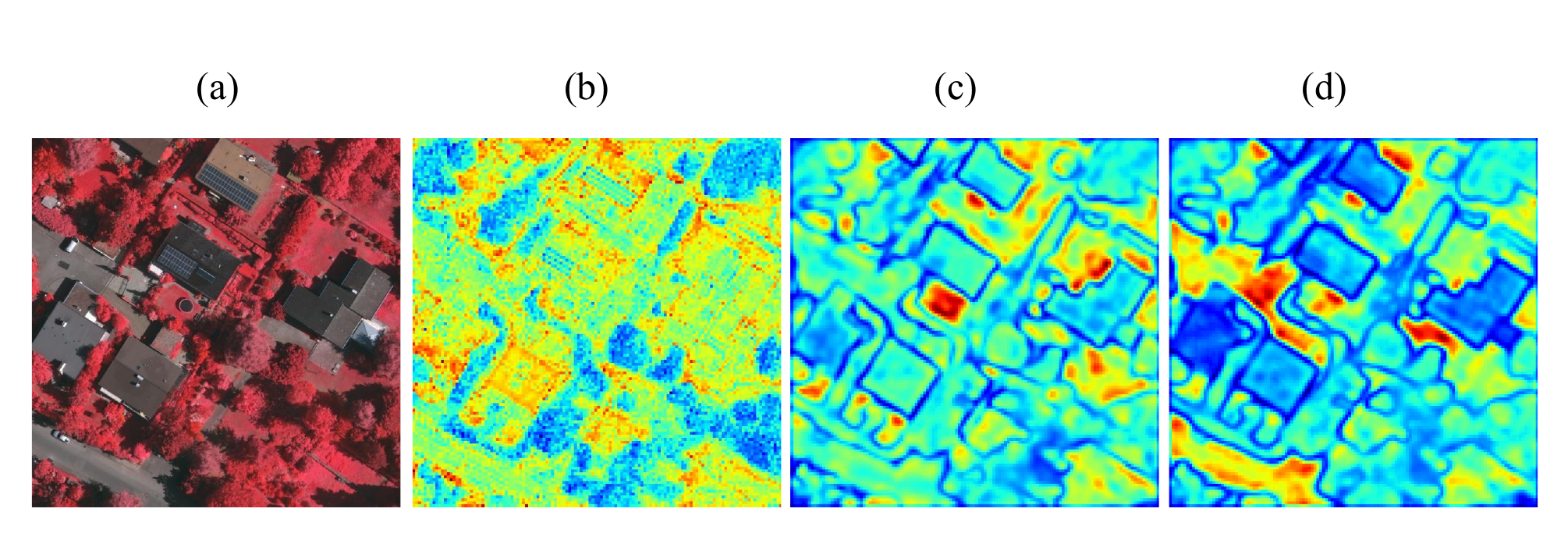}
			\caption{Feature maps before and after using APEM and SSCM. (a) Remote sensing image. (b) Feature map without using APEM and SSCM. (c) Feature map after applying APEM. (d) Feature map after applying APEM and SSCM. 
			}
			\label{fig8}
		\end{figure}

			%

		\begin{table}[htbp]
			\centering
			\setlength{\tabcolsep}{7.3pt} 
			\caption{Results  of  pdssnet with individual componets removed. The best results are indicated in bold black.
				\label{tab1}}
			\begin{tabular}{l|cc|cc}
				\hline
				
				\multirow{2}{*}[-0.1ex]{\makecell[c]{\hspace{20pt}Config\bigstrut[t]}} & \multicolumn{2}{c|}{Vaihingen\bigstrut[t]} & \multicolumn{2}{c}{Potsdam\bigstrut[t]} \\
				\cline{2-3} \cline{4-5}
				& {mIoU(\%)\bigstrut[t]} & {F1(\%)\bigstrut[t]} & {mIoU(\%)\bigstrut[t]} & {F1(\%)\bigstrut[t]} \\
				
				\hline
				
				PDSSNet w/o APEM & 83.75 & 91.04 & 86.98 & 92.92 \bigstrut[t]\\
				PDSSNet w/o SSCM & 83.81 & 91.08 & 86.76 & 92.80 \bigstrut[t]\\
				PDSSNet w/o CSAM & 84.20 & 91.32 & 87.07 & 92.98 \bigstrut[t]\\
				PDSSNet w/o \(\mathcal{L}_{ce}\)& 84.06 & 91.23 & 87.02 & 92.96  \bigstrut[t]\\
				PDSSNet w/o \(\mathcal{L}_{dice}\)& 84.36 & 91.41 & 87.30 & 93.11  \bigstrut[t]\\
				PDSSNet & \textbf{84.68} & \textbf{91.60} & \textbf{87.55} & \textbf{93.25} \bigstrut[t]\\
				
				\hline
				
			\end{tabular}
		\end{table}

		\subsubsection{The impact of APEM}
		As shown in the data from Table \ref{tab1}, removing the APEM from the PDSSNet model leads to a 0.93\% decrease in mIoU and a 0.56\% decrease in F1 on the Vaihingen dataset. On the Potsdam dataset, the corresponding mIoU decreases by 0.57\% and the F1 decreases by 0.33\%. Conversely, when the APEM is added to the baseline model, the mIoU increases by 1.09\% and the F1 increases by 0.67\% on the Vaihingen dataset. On the Potsdam dataset, the mIoU increases by 1.14\% and the F1 increases by 0.66\%.
		The visualization results in Fig. \ref{fig7} visually demonstrate the effectiveness of APEM in addressing high intra-class variance. After incorporating APEM into the baseline model, the first row shows that for scenarios where the same class exhibits different manifestations, APEM effectively reduces class mis-predictions. The second row indicates that for the issue of two cars being mistakenly segmented as one due to shadows, adding APEM achieves clear class separation. This fully demonstrates that by constructing comprehensive class prototypes from GT, APEM significantly enhances both class representation and intra-class feature aggregation capabilities.
		Further analysis of the visualization results in Fig. \ref{fig6} shows that when APEM is removed from PDSSNet, the integrity of segmentation results is significantly compromised, with issues such as fragmentation and boundary blurring appearing in target areas. Fig. \ref{fig8} displays the feature maps before and after the application of APEM. It can be observed that with the use of APEM, the contours of the ground objects become substantially clearer, and the overall color contrast is significantly enhanced.
		To validate the rationality of the selected update parameter, comparative experiments with different parameter settings were conducted, and the results are shown in Table \ref{tab8}. The experimental results demonstrate that setting \(\bm{\beta}\) to 0.7 yields the optimal value.

		\begin{figure}[!t]
			\centering
			\includegraphics[width=3.5in]{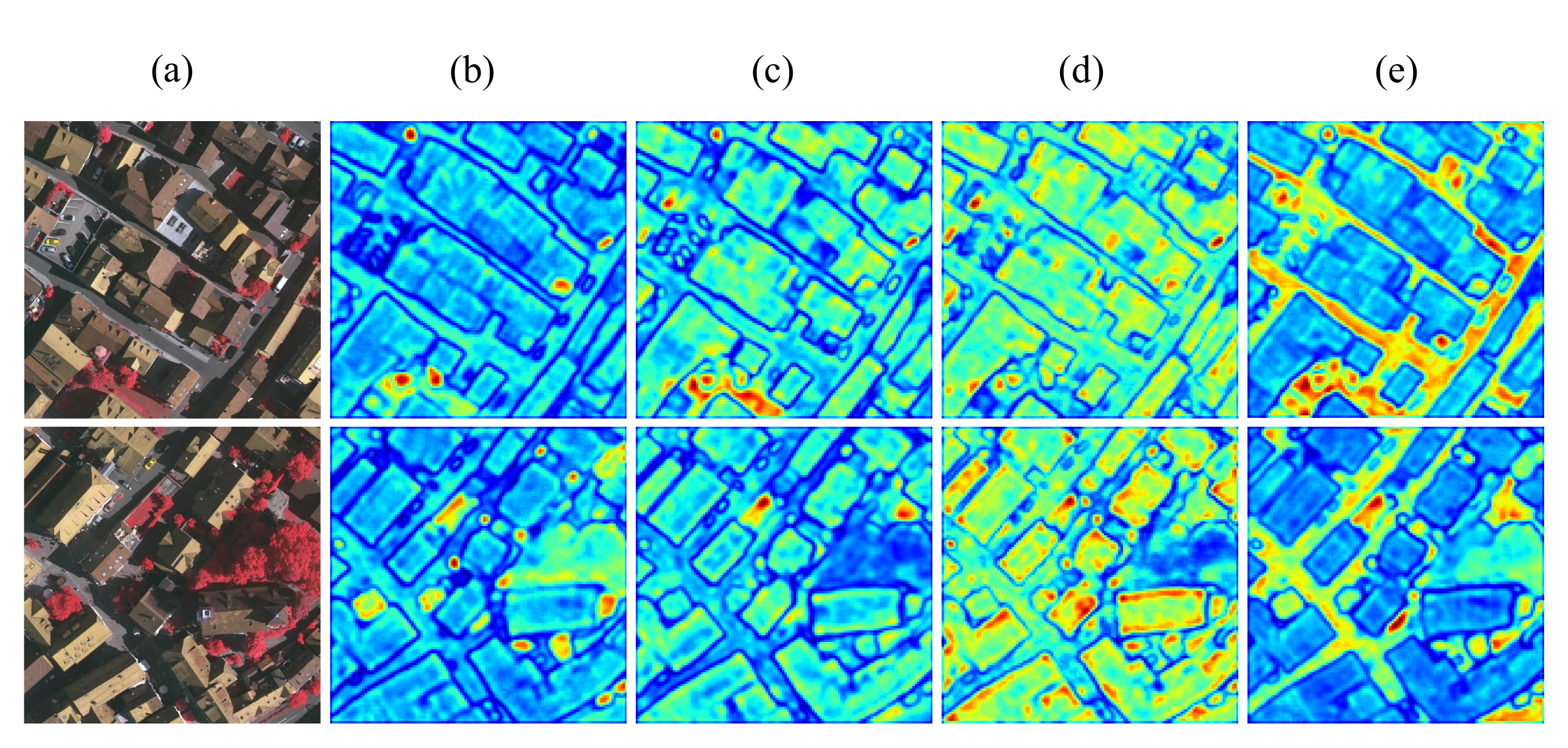}
			\caption{Feature visualization of three feature interaction methods is provided. (a) Remote sensing image. (b) Category emphasis method. (c) Self-attention mechanism. (d) VSS method. (e) Using SSCM method.
			}
			\label{fig9}
		\end{figure}

		\begin{table}[htbp]
			\centering
			\setlength{\tabcolsep}{9pt}
			\caption{Ablation study of  pdssnet with different component combinations.	\label{tab7}}
			\begin{tabular}{l|cc|cc}
				\hline
				\multirow{2}{*}[-0.1ex]{\makecell[c]{\hspace{20pt}Config\bigstrut[t]}} & \multicolumn{2}{c|}{Vaihingen\bigstrut[t]} & \multicolumn{2}{c}{Potsdam\bigstrut[t]} \\
				\cline{2-3} \cline{4-5}
				& {mIoU(\%)\bigstrut[t]} & {F1(\%)\bigstrut[t]} & {mIoU(\%)\bigstrut[t]} & {F1(\%)\bigstrut[t]} \\
				
				\hline
				
				Baseline & 82.60 & 90.34 & 85.39 & 92.00\bigstrut[t] \\
				Baseline + APEM & 83.69 & 91.01 & 86.53 & 92.66\bigstrut[t] \\
				Baseline + SSCM & 83.73 & 91.03 & 86.68 & 92.75\bigstrut[t] \\
				Baseline + CSAM & 83.53 & 90.91 & 86.54& 92.67\bigstrut[t] \\
				
				\hline
				
			\end{tabular}
		\end{table}

		\subsubsection{The impact of SSCM}
		As indicated by the data in Table \ref{tab1}, removing the SSCM from the PDSSNet model causes a 0.87\% decrease in mIoU and a 0.52\% decrease in F1 on the Vaihingen dataset. On the Potsdam dataset, the corresponding mIoU decreases by 0.79\% and F1 decreases by 0.45\%. Conversely, when the SSCM is added to the baseline model, the mIoU increases by 1.13\% and F1 increases by 0.69\% on the Vaihingen dataset. On the Potsdam dataset, the mIoU increases by 1.15\% and F1 increases by 0.67\%. These experimental results collectively demonstrate that the SSCM can improve mIoU by at least 0.76\% and F1 by at least 0.45\%.
		As the visualization results in Fig. \ref{fig7} demonstrate, integrating the SSCM further enhances segmentation integrity. This is evidenced by the reduction in erroneous pixel predictions in the first row and the clearer separation of adhered vehicles in the second. This is because the module first utilizes the high-quality prototypes from APEM to establish a global semantic understanding. It then leverages structural information to constrain and refine this initial semantic representation. This refinement process corrects initial misjudgments and ensures the final segmentation result conforms to the true object morphology, thereby significantly improving the completeness of the result.

		Fig. \ref{fig8} visually illustrates the refinement process of feature representations by APEM and SSCM. Initially, in the baseline's original feature map, the activation regions are rather diffuse, and object contours are indistinct. After introducing the high-quality semantic priors from APEM, the basic contours of ground objects (especially buildings) begin to emerge, and the overall contrast of the feature map is enhanced. The addition of SSCM, however, brings a qualitative leap. Not only do the object contours become extremely clear and sharp, but the intra-class feature responses also become more consistent and intense, and the separability from the background is significantly improved.

		\begin{table}[htbp]
			\centering
			\setlength{\tabcolsep}{16pt}
			\caption{Comparison of different feature-guided.}
			\label{tab:Comparison of different feature-guided }\label{tab2}
			\begin{tabular}{l|cc}
				
				\hline
				\multirow{2}{*}[-0.1ex]{\makecell[c]{\hspace{30pt}Config\bigstrut[t]}} &  
				\multicolumn{2}{c}{Vaihingen\bigstrut[t]}  \\
				\cline{2-3} 
				&{mIoU(\%)\bigstrut[t]} & {F1(\%)\bigstrut[t]}  \\
				\hline
				PDSSNet w/o SSCM + CW  & 84.03 & 91.21 \bigstrut[t] \\
				PDSSNet w/o SSCM + SA & 84.24 & 91.34 \bigstrut[t] \\
				PDSSNet w/o F-SS2D + VSS & 84.36 &91.41\bigstrut[t]\\
				PDSSNet  & \textbf{84.68} & \textbf{91.60} \bigstrut[t] \\
				\hline
			\end{tabular}
		\end{table}
		
		\begin{table}[htbp]
			\centering
			\setlength{\tabcolsep}{12.5pt}
			\caption{Comparison of different backbones.}
			\label{tab3}
			\begin{tabular}{l|c|c|c}
				\hline
				Method\bigstrut[t] & Backbone\bigstrut[t] & Params(M)\bigstrut[t] & mIoU(\%)\bigstrut[t] \\
				\hline
				\multirow{8}{*}{PDSSNet\bigstrut[t]} 
				& ResNet50 & 25.56 & 84.17\bigstrut[t] \\
				& ResNext50 & 25.03 & 84.32\bigstrut[t] \\
				& ResNest50 & 25.48 & 84.29 \bigstrut[t]\\
				& Maxvit-Tiny & 31.05 & 84.44\bigstrut[t] \\
				& Swinv2-Tiny & 28.33 & 84.54 \bigstrut[t]\\
				& Vim-Small & 25.87 & 84.41 \bigstrut[t]\\
				& Vmamba-Tiny & \textbf{22.37} & 84.52\bigstrut[t] \\
				& ConvNext-Tiny & 28.59 & \textbf{84.68} \bigstrut[t]\\
				\hline
			\end{tabular}
		\end{table}




		\begin{figure*}[h]
			\centering
			\includegraphics[width=7.25in]{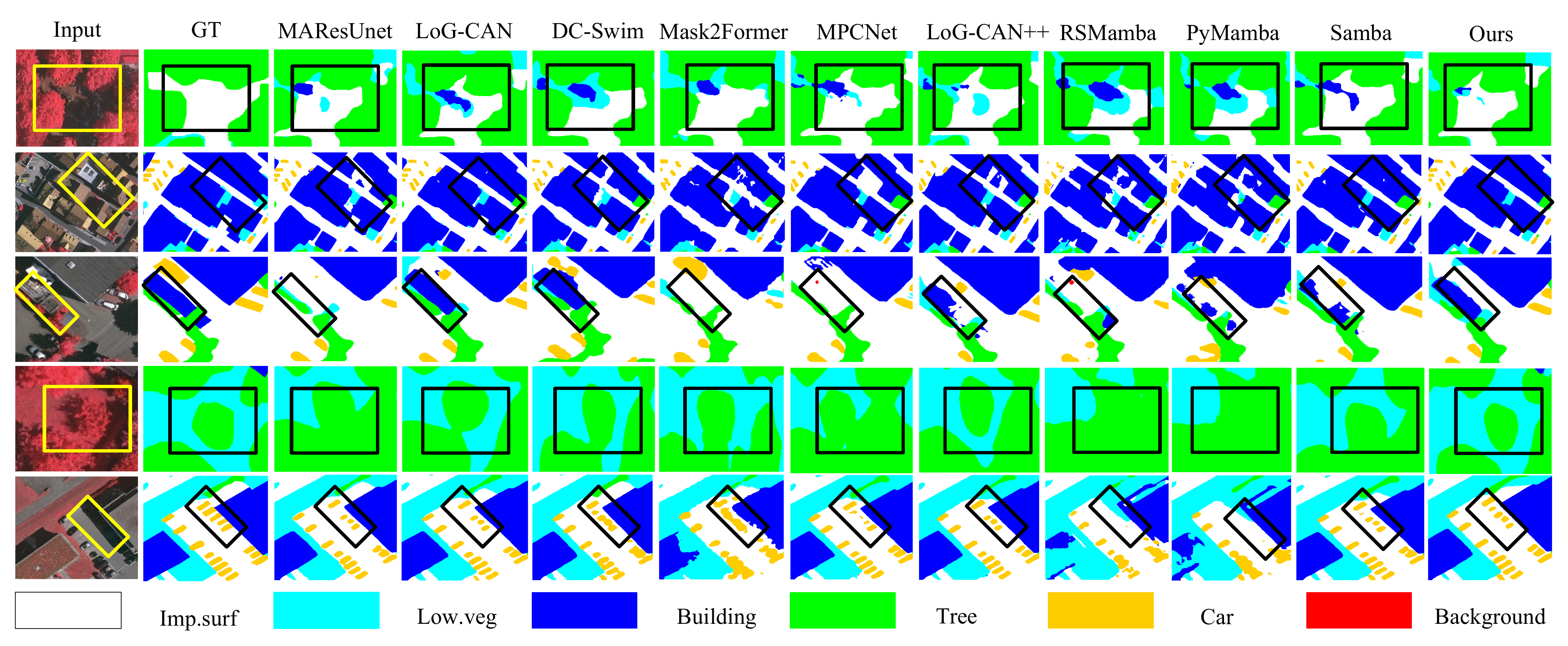}
			\caption{A local zoomed view of the segmentation results on the Vaihingen dataset of PDSSNet and other comparison models.}
			\label{fig12}
		\end{figure*}

		\begin{figure*}[h]
			\centering
			\includegraphics[width=7.2in]{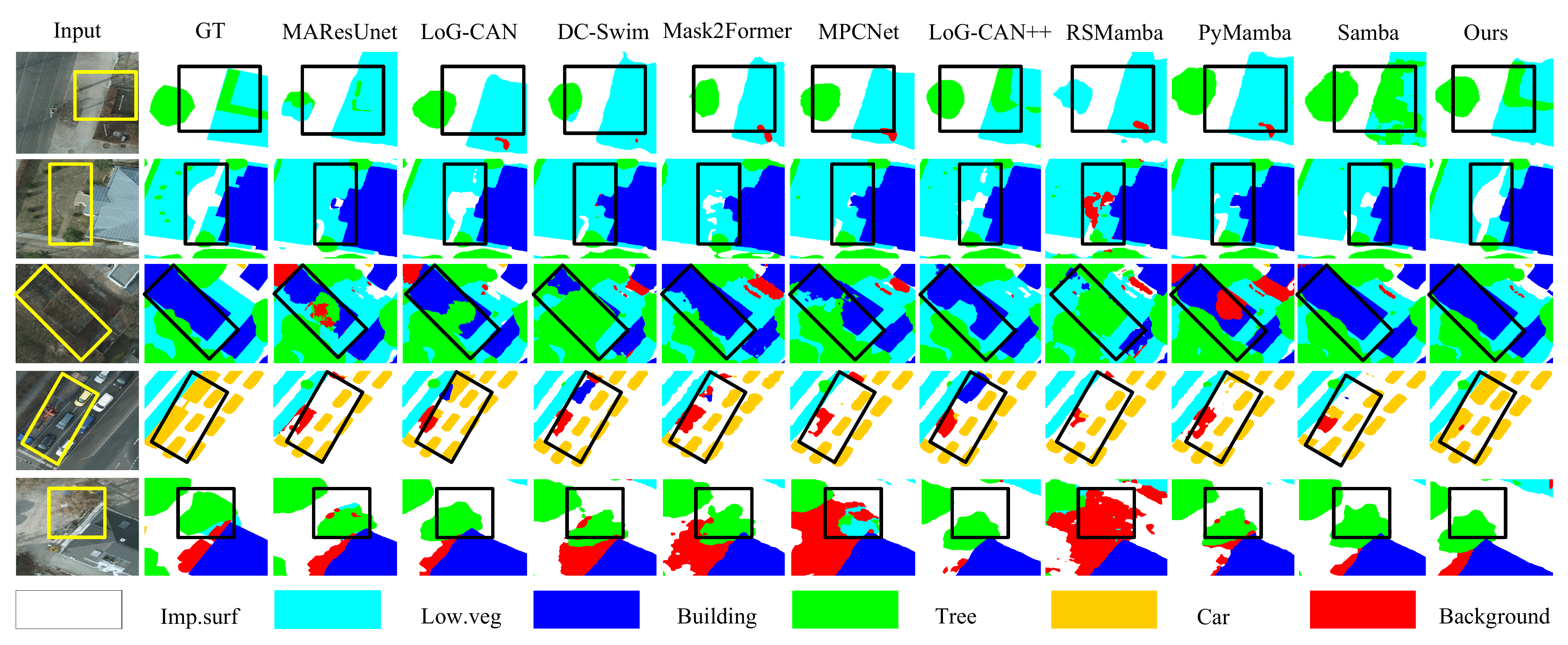}
			\caption{A local zoomed view of the segmentation results on the Potsdam dataset of PDSSNet and other comparison models.}
			\label{fig14}
		\end{figure*}

		\begin{figure*}[h]
			\centering
			\includegraphics[width=7.3in]{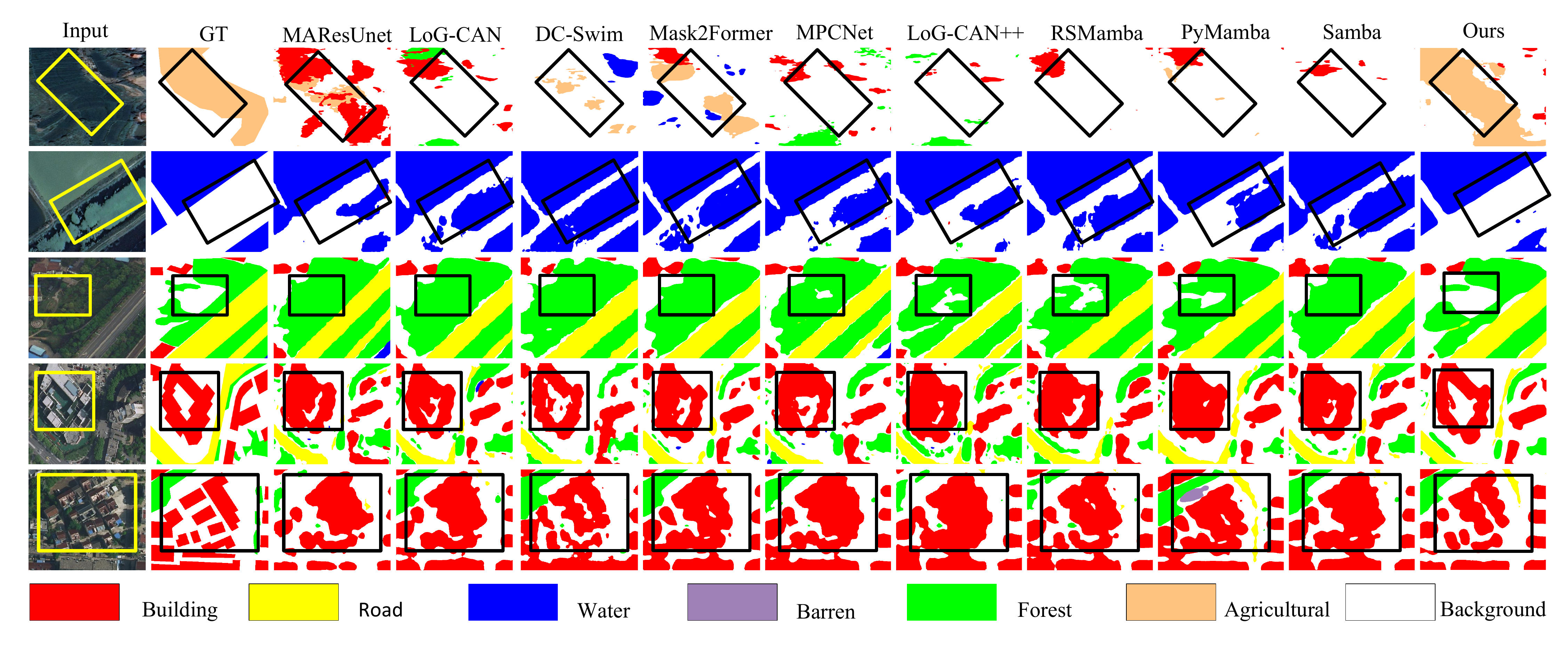}
			\caption{A local zoomed view of the segmentation results on the LoveDA dataset of PDSSNet and other comparison models.}
			\label{fig15}
		\end{figure*}
		
		\begin{table*}[!ht]
			\centering
			\setlength{\tabcolsep}{13.5pt}
			\caption{Comparison of segmentation results on the vaihingen dataset.
			}
			
			\label{tab4}
			\begin{tabular}{l|ccccc|ccc}
				
				\hline
				\multirow{2}{*}[-0.1ex]{\makecell[c]{\hspace{20pt}Method}\bigstrut[t]} & \multicolumn{5}{c|}{F1(\%)}\bigstrut[t] & \multicolumn{3}{c}{Evaluation index}\bigstrut[t] \\
				\cline{2-6} \cline{7-9}
				& {Imp.Surf}\bigstrut[t] &{Building}\bigstrut[t] & {Lowveg}\bigstrut[t] & {Tree}\bigstrut[t] & {Car}\bigstrut[t] & {MeanF1(\%)}\bigstrut[t] & {mIoU(\%)}\bigstrut[t] & {OA(\%)}\bigstrut[t] \\
				\hline
				MAResU-Net\cite{58} & 92.91 & 95.26 & 84.95 & 89.94 & 88.33 & 90.28 & 83.30 & 90.86 \bigstrut[t]\\
				LoG-CAN\cite{32} & 93.00 & 95.67 & 84.86 & 90.28 & 88.27 & 90.42 & 82.72 & 91.20 \bigstrut[t]\\
				DC-Swin\cite{60} & 93.46 & 96.00 & 85.32 & 90.03 & 84.88 & 89.94 & 82.01 & 91.29 \bigstrut[t]\\
				Mask2Former\cite{61} & 92.86 & 96.03 & 84.15 & 90.50 & 89.30 & 90.57 & 82.99 & 91.10 \bigstrut[t]\\
				MPCNet\cite{62} & 92.76 & 95.50 & 84.70 & 90.40 & 90.44 & 90.76 & 83.27 & 90.93 \bigstrut[t]\\
				LoG-CAN++\cite{88} & 93.41 & 95.93 & 85.26 & 90.59 & 89.46 & 90.93 & 83.57 & 91.50 \bigstrut[t]\\
				RS-Mamba\cite{28} & 93.10 & 95.61 & 84.09 & 90.26 & 89.42 & 90.50 & 82.87 & 90.98 \bigstrut[t]\\
				PyMamba\cite{27} & 92.57 & 95.41 & 84.73 & 90.53 & 88.27 & 90.30 & 82.52 & 90.91 \bigstrut[t]\\
				Samba\cite{30} & 93.03 & 95.44 & 83.84 & 90.70 & 89.11 & 90.42 & 82.75 & 90.99 \bigstrut[t]\\
				\hline
				PDSSNet(Ours) & \textbf{93.46} & \textbf{96.28} & \textbf{86.38} & \textbf{91.39} & \textbf{90.50} & \textbf{91.60} & \textbf{84.68} & \textbf{91.94}\bigstrut[t] \\
				\hline
				
			\end{tabular}
		\end{table*}

		To validate the effectiveness of the core algorithm F-SS2D compared with category-guided interaction methods, we conducted replacement experiments in PDSSNet: SSCM was substituted with common methods such as self-attention mechanisms and category weighting (as shown in Fig. \ref{fig3}), and compared with the similar method SS2D. The relevant experimental results are shown in Table \ref{tab2} (CW denotes the class weight method, SA denotes the self-attention method, and VSS denotes the original VMamba method) and Fig. \ref{fig9}. 
		The visualization results of the latter indicate that after processing with F-SS2D, the color contrast of the features is significantly enhanced, and there is a greater focus on ground objects that are easily confused or difficult to identify. Specifically, for large regular ground objects, the APEM stage can already outline their general contours through semantic information. For difficult-to-identify ground objects, the first stage performs preliminary positioning based on category semantic information, and the second stage conducts fine-grained adjustment through spatial structure constraints. This result verifies the advantages of the proposed hierarchical cognitive strategy and highlights its advancement in improving the segmentation accuracy of remote sensing images.

		\subsubsection{The impact of CSAM}
		As shown in the data from Table \ref{tab1}, removing the CSAM from the PDSSNet model leads to a 0.48\% decrease in mIoU and a 0.28\% decrease in F1 on the Vaihingen dataset. On the Potsdam dataset, the corresponding mIoU and F1 decrease by 0.48\% and 0.27\%, respectively. Conversely, when the CSAM is added to the baseline model, the mIoU increases by 0.93\% and the F1 increases by 0.57\% on the Vaihingen dataset. On the Potsdam dataset, the mIoU and F1 increase by 1.29\% and 0.75\%, respectively. The experimental results collectively demonstrate that the CSAM can effectively enhance model performance and plays a crucial role in the model's architecture.
		
		As shown in the visualization results of Fig. \ref{fig6}, after removing the CSAM, the model exhibited significant mispredictions when segmenting easily confused categories.
		As shown in the visualization results of Fig. \ref{fig7}, after incorporating CSAM into the baseline model, the first row demonstrates complete segmentation of category information, and the adhesion between cars in the second row is significantly alleviated. This is because when CSAM encounters similar features, it breaks free from local similarity constraints and infers the category of similar features from other regions, thereby augmenting the model’s competence to discriminate confusing features.

		\subsubsection{Impact of the Loss Function}
		As observed from the experimental results in Table \ref{tab1}, different configurations of loss functions have varying degrees of impact on model performance. On the Vaihingen dataset, removing    \(\mathcal{L}_{ce}\) causes mIoU to decrease by 0.62\% and F1 by 0.37\%, while removing \(\mathcal{L}_{dice}\) loss leads to a 0.32\% decrease in mIoU and a 0.19\% decrease in F1. On the Potsdam dataset, removing \(\mathcal{L}_{ce}\) results in a 0.53\% drop in mIoU and a 0.30\% drop in F1, whereas removing \(\mathcal{L}_{dice}\) has a smaller effect, with mIoU decreasing by 0.25\% and F1 by 0.14\%.
		Removing \(\mathcal{L}_{ce}\), which is used to optimize pixel-level classification and boundary distinction, significantly reduces the mIoU and F1 scores. In contrast, removing \(\mathcal{L}_{dice}\), which primarily addresses the issues of region overlap and class imbalance, has a lesser impact, indicating that its role is more auxiliary.
		Overall, combining \(\mathcal{L}_{ce}\) and  \(\mathcal{L}_{dice}\) balances the needs of inter-class discrimination and regional overlap optimization. Removing any of these loss functions will negatively affect model performance to some extent.

		\subsubsection{The impact of Backbone}
				To reduce the backbone's interference with experimental results, we carried out backbone replacement experiments on the Vaihingen dataset. The experiments selected multiple widely used backbone architectures with similar parameter scales, including convolutional networks (such as ConvNext-Tiny, ResNet 50, ResNext 50, and ResNest 50), Transformer-based networks (such as Maxvit-Tiny\cite{57} and Swinv2\cite{56} Tiny), and Mamba-based structures (such as Vim-Small\cite{55} and Vmamba-Tiny\cite{26}). The experimental results are shown in Table \ref{tab3}, where all backbones exhibit high performance metrics. In terms of the number of parameters, Vmamba-Tiny has the fewest parameters in all experiments, demonstrating a significant computational efficiency advantage. However, when comprehensively balancing the number of parameters and accuracy, ConvNext-Tiny shows the best performance. It can achieve high accuracy while maintaining a low number of parameters. Based on this, we selected it as our backbone.


		\begin{table*}[!ht]
			\centering
			\setlength{\tabcolsep}{13.5pt}
			\caption{Comparison of segmentation results on the potsdam dataset.}
			\label{tab5}
			\begin{tabular}{l|ccccc|ccc}
				\hline
				\multirow{2}{*}[-0.1ex]{\makecell[c]{\hspace{20pt}Method}\bigstrut[t]} & \multicolumn{5}{c|}{F1(\%)\bigstrut[t]} & \multicolumn{3}{c}{Evaluation index}\bigstrut[t] \\
				\cline{2-6} \cline{7-9}
				& {Imp.Surf}\bigstrut[t] & {Building}\bigstrut[t] & {Lowveg}\bigstrut[t] & {Tree}\bigstrut[t] & {Car}\bigstrut[t] & {MeanF1(\%)}\bigstrut[t] & {mIoU(\%)}\bigstrut[t] & {OA(\%)}\bigstrut[t] \\
				\hline
				MAResU-Net\cite{58} & 92.38 & 95.83 & 86.65 & 88.44 & 96.13 & 91.89 & 85.22 & 90.28 \bigstrut[t]\\
				LoG-CAN\cite{33} & 93.36 & 96.80 & 87.63 & 88.83 & 95.87 & 92.50 & 86.26 & 91.22 \bigstrut[t]\\
				DC-Swin\cite{60} & 93.26 & 96.86 & 87.74 & 88.68 & 95.50 & 92.41 & 86.10 & 91.16 \bigstrut[t]\\
				Mask2Former\cite{61} & 92.48 & 96.41 & 87.53 & 89.37 & 96.08 & 92.37 & 86.03 & 90.83 \bigstrut[t]\\
				MPCNet\cite{62} & 92.69 & 96.38 & 87.30 & 88.74 & 96.34 & 92.29 & 85.91 & 90.56 \bigstrut[t]\\
				LoG-CAN++\cite{88} & \textbf{93.82} & 97.12 & \textbf{88.73} & 89.60 & 96.83 & 93.22 & 87.50 & 91.78 \bigstrut[t]\\
				RS-Mamba\cite{28} & 92.56 & 96.63 & 87.43 & 88.51 & 95.72 & 92.17 & 85.70 & 90.70 \bigstrut[t]\\
				PyMamba\cite{27} & 92.33 & 96.32 & 86.99 & 88.62 & 96.23 & 92.10 & 85.59 & 90.48 \bigstrut[t]\\
				Samba\cite{30} & 92.87 & 96.99 & 87.49 & 89.16 & 95.74 & 92.45 & 86.17 & 91.05 \bigstrut[t]\\
				\hline
				PDSSNet(Ours) & 93.80 & \textbf{97.30} & 88.45 & \textbf{89.97} & \textbf{96.70} & \textbf{93.25} & \textbf{87.55} & \textbf{91.84} \bigstrut[t]\\
				\hline
				
			\end{tabular}
		\end{table*}
		
		\begin{table*}[!ht]
			\centering
			\setlength{\tabcolsep}{8pt}
			\caption{Comparison of segmentation results on the loveda dataset.}
			\label{tab6}
			\begin{tabular}{l|ccccccc|ccc}
				\hline
				\multirow{2}{*}[-0.1ex]{\makecell[c]{\hspace{20pt}Method}}\bigstrut[t] & \multicolumn{7}{c|}{F1(\%)} \bigstrut[t]& \multicolumn{3}{c}{Evaluation index}\bigstrut[t] \\
				\cline{2-8} \cline{9-11}
				& {Building}\bigstrut[t]  &{Road}\bigstrut[t]  &{Water} \bigstrut[t] &{Barren}\bigstrut[t]& {Forest}\bigstrut[t] & {Agricultural}\bigstrut[t] & {Background}\bigstrut[t] & {MeanF1(\%)}\bigstrut[t] & {mIoU(\%)}\bigstrut[t] & {OA(\%)}\bigstrut[t] \\
				\hline
				MAResU-Net\cite{58} & 77.75 & 70.86 & 77.10 & 38.59 & 61.05 & 68.58 & 68.34 & 66.04 & 50.45 & 68.92 \bigstrut[t]\\
				LoG-CAN\cite{33} & 68.74 & 75.96 & 69.14 & \textbf{80.05} & 41.08 & 60.69 & 68.62 & 65.94 & 50.43 & 69.12 \bigstrut[t]\\
				DC-Swin\cite{60} & 72.71 & 72.28 & 83.30 & 49.23 & 57.42 & 62.17 & 69.00 & 66.59 & 50.83 & 68.01 \bigstrut[t]\\
				Mask2Former\cite{61} & 77.51 & 72.59 & 83.34 & 44.07 & 60.97 & 68.16 & 70.93 & 68.23 & 52.92 & 70.82 \bigstrut[t]\\
				MPCNet\cite{62} & 78.82 & 71.16 & 83.45 & 46.44 & 55.55 & 70.38 & 70.76 & 68.08 & 52.80 & 70.87 \bigstrut[t]\\
				LoG-CAN++\cite{88} & 70.99 & {77.30} & {71.24} & 79.79 & 48.36 & 58.37 & 68.05 & 67.68 & 52.14 & 70.22 \bigstrut[t]\\
				RS-Mamba\cite{28} & 78.79 & 72.50 & 83.21 & 43.61 & \textbf{61.52} & 69.41 & 70.88 & 68.56 & 53.35 & 71.16 \bigstrut[t]\\
				PyMamba\cite{27} & 76.76 & 66.56 & 82.28 & 53.52 & 54.60 & 66.81 & 69.68 & 67.17 & 51.40 & 69.10 \bigstrut[t]\\
				Samba\cite{30} & 76.01 & 67.84 & 81.98 & 46.37 & 59.43 & 66.98 & 69.61 & 66.89 & 51.19 & 69.19 \bigstrut[t]\\
				\hline
				PDSSNet(Ours) & 78.82 &\textbf {73.52} & \textbf {85.13} & 55.94 & 60.18 & \textbf{71.68} & \textbf{73.14} & \textbf{71.20} & \textbf{56.10} & \textbf{73.02} \bigstrut[t]\\
				\hline
				
			\end{tabular}
		\end{table*}
		
		\begin{table}[htbp]
			\centering
			\setlength{\tabcolsep}{13.5pt}
			\caption{Different parameter settings. 	\label{tab8}}
			\begin{tabular}{l|cc|cc}
				\hline
				\multirow{2}{*}[-0.1ex]{\makecell[c]{\hspace{4pt}\(\bm{\beta}\)}}\bigstrut[t] & \multicolumn{2}{c|}{Vaihingen}\bigstrut[t] & \multicolumn{2}{c}{Potsdam}\bigstrut[t] \\
				\cline{2-3} \cline{4-5}
				& {mIoU(\%)} \bigstrut[t]& {F1(\%)}\bigstrut[t] & {mIoU(\%)}\bigstrut[t] & {F1(\%)}\bigstrut[t] \\
				\hline
				0.8 & 84.48 & 91.49 & 87.40 & 93.16\bigstrut[t] \\
				0.7  & \textbf{84.68} & \textbf{91.60} & \textbf{87.55} & \textbf{93.25}\bigstrut[t] \\
				0.6  & 84.63 & 91.57 & 87.20 & 93.04\bigstrut[t] \\
				0.5  & 84.65 & 91.59 & 87.19 & 93.04 \bigstrut[t]\\
				
				\hline
			\end{tabular}
		\end{table}

		\subsection{Comparative Experiments}
		When conducting model comparisons, we selected multiple models as comparison objects, including the Multi-level Attention Residual UNet (MAResU-Net\cite{58}) using linear attention mechanisms, the Swin-Transformer network with a Dense Connection Feature Aggregation Module (DCFAM) (DC-Swin\cite{60}), the Transformer network Mask2Former\cite{61} using masks for attention calculation, the network with a multi-scale prototype-based Transformer decoder (MPCNet\cite{62}), LoG-CAN\cite{33} and LoG-CAN++\cite{88}, which also adopt category-guided methods, RS-Mamba\cite{28}, PyramidManba\cite{27}, and Samba\cite{30} also use  SSM for dense prediction tasks.
		PDSSNet outperformed the aforementioned models in accuracy on the ISPRS Vaihingen, ISPRS Potsdam, and LoveDA datasets, which are commonly employed for remote sensing segmentation research.  
		Tables \ref{tab4}-\ref{tab6} present the comparison results between PDSSNet and other models on the Vaihingen dataset, Potsdam dataset, and LoveDA dataset, respectively.

		\subsubsection{Results of the Vaihingen Dataset}

		In the experiments on the Vaihingen dataset, as presented in Table \ref{tab4}, PDSSNet achieved significant improvements in mean F1, mean Intersection over Union (mIoU), and Overall Accuracy (OA), reaching 91.60\%, 84.68\%, and 91.94\%, respectively.
		Specifically, PDSSNet outperforms pure convolutional attention-based networks such as MAResU-Net, as well as category-guided method networks like LoG-CAN and LoG-CAN++. Compared with RS-Mamba, which also uses the SSM, PDSSNet's F1 is 1.10\% higher, mIoU is 1.81\% higher, and OA is 0.96\% higher, fully demonstrating the excellent segmentation performance of PDSSNet. The effect is shown in Fig. \ref{fig12}. 
		
		This quantitative performance advantage is primarily attributed to PDSSNet's exceptional ability to address the two core challenges of remote sensing imagery. In handling high intra-class variance (as shown in the second row of Fig. \ref{fig12}), PDSSNet maintains the structural integrity of buildings even with variations in lighting and material on rooftops, whereas other methods (like Mask2Former and LoG-CAN) exhibit noticeable internal holes and fragmentation. This directly reflects the synergistic advantage of APEM and SSCM.
		In dealing with highly similar classes (as shown in the fifth  rows of Fig. \ref{fig12}), our model achieves cleaner segmentation when vehicles are parked closely together or confused with shadows, effectively mitigating the adhesion problems found in other methods. This highlights the effectiveness of the CSAM.
		
		
		\subsubsection{Results of the Potsdam Dataset}
		To comprehensively evaluate the segmentation capabilities of PDSSNet, we conducted comparative experiments on the Potsdam dataset. As shown in Table \ref{tab5}, PDSSNet achieved an F1 score, mIoU, and OA of 93.25\%, 87.55\%, and 91.84\%, respectively, on the Potsdam test set, with all metrics significantly surpassing those of competing methods.
		Compared to RS-Mamba, which also employs a SSM, PDSSNet demonstrated a lead of 1.04\%, 1.76\%, and 1.06\% in F1, mIoU, and OA, respectively.Table \ref{tab5} shows the prediction outcomes for the various semantic segmentation methods. The experimental results indicate that the prototype-driven structural constraint method exhibits a significant advantage in the complete segmentation of ground objects.
		
		This quantitative performance advantage is primarily attributed to PDSSNet's outstanding performance in addressing the core challenge of high inter-class similarity, a point fully corroborated by the visualization results in Fig. \ref{fig14}. For example, in the first row of Fig. \ref{fig14}, our model is still able to accurately delineate the clear boundaries of a building even when its rooftop shares highly similar spectral features with the adjacent impervious surface, whereas other comparative methods produce noticeable class confusion and erroneous segmentation.
		This powerful discriminative capability stems from the backward update of class prototypes by the structure-refined features, which enables the prototypes to learn more complex, context-aware representations. Furthermore, the SimStep mechanism designed within the CSAM module allows the model to “skip” local, easily confusable similar regions and instead focus on more discriminative boundary information. It is precisely this effective modeling of both high intra-class variance and high inter-class similarity that enables our model to perform exceptionally well on typical ground object classes such as buildings and cars.
		
		\subsubsection{Results of the LoveDA Dataset}
		For additional verification of PDSSNet’s generalization ability, comparative experiments were conducted on the LoveDA remote sensing dataset. The results in Table \ref{tab6} show that PDSSNet significantly outperforms other compared networks in terms of average F1, mIoU, and OA on this dataset, reaching 71.20\%, 56.10\%, and 73.02\%, respectively, with outstanding overall performance.Notably, PDSSNet significantly outperforms other networks in segmenting agricultural and water categories. This is because agricultural areas in rural scenes usually exhibit regular shapes, making it easier for the model to learn the representation of this category after determining semantic information. For water categories, urban water bodies often feature regularly shaped artificial designs, while rural water areas typically distribute along or within the boundaries of agricultural regions. The semantic stage of SSCM helps the model infer the spatial distribution characteristics of rural water areas by establishing spatial dependency relationships between categories, thus enabling better segmentation performance for these two types of ground objects compared to other networks.
		
		Additionally, we visualized and compared the prediction results of multiple semantic segmentation methods mentioned in Table \ref{tab6}. As shown in Fig. \ref{fig15}, when facing the challenges of multi-scale objects intertwined with complex backgrounds in the LoveDA dataset, PDSSNet demonstrates significant advantages in complex ground object segmentation tasks by virtue of its precise analysis capability of scene features. Compared with other methods, the model can outline target boundaries more clearly, effectively reduce category confusion, and greatly improve the integrity and accuracy of segmentation results, fully verifying its excellent adaptability and leading performance in complex remote sensing scenarios.

		\section{CONCLUSION}
		To address the problem of incomplete class segmentation in remote sensing images, this paper proposed the Prototype-Driven Structure-Synergized Network (PDSSNet). Our proposed APEM established a new method for class modeling. By leveraging ground truth, it constructs comprehensive class prototypes, thereby effectively overcoming the representational limitations of traditional methods when handling high intra-class variance. The SSCM module employs a hierarchical “semantic judgment-structural constraint” cognitive strategy, with the core objective of driving the dynamic alignment of high-level semantics and spatial structure to ensure the morphological integrity of the segmentation results. Finally, the CSAM adopts a novel SimStep mechanism, effectively enhancing the model's ability to distinguish between confusable classes by adaptively adjusting its focus.
		Experiments validated the overall effectiveness of PDSSNet as well as the performance of its individual modules. We hope that PDSSNet will offer valuable insights to future researchers working to solve the problem of incomplete class segmentation.
		
		

		\bibliographystyle{IEEEtran}
		\bibliography{references}{}
		\begin{IEEEbiography}[{\includegraphics[width=1in,height=1.25in,clip,keepaspectratio]{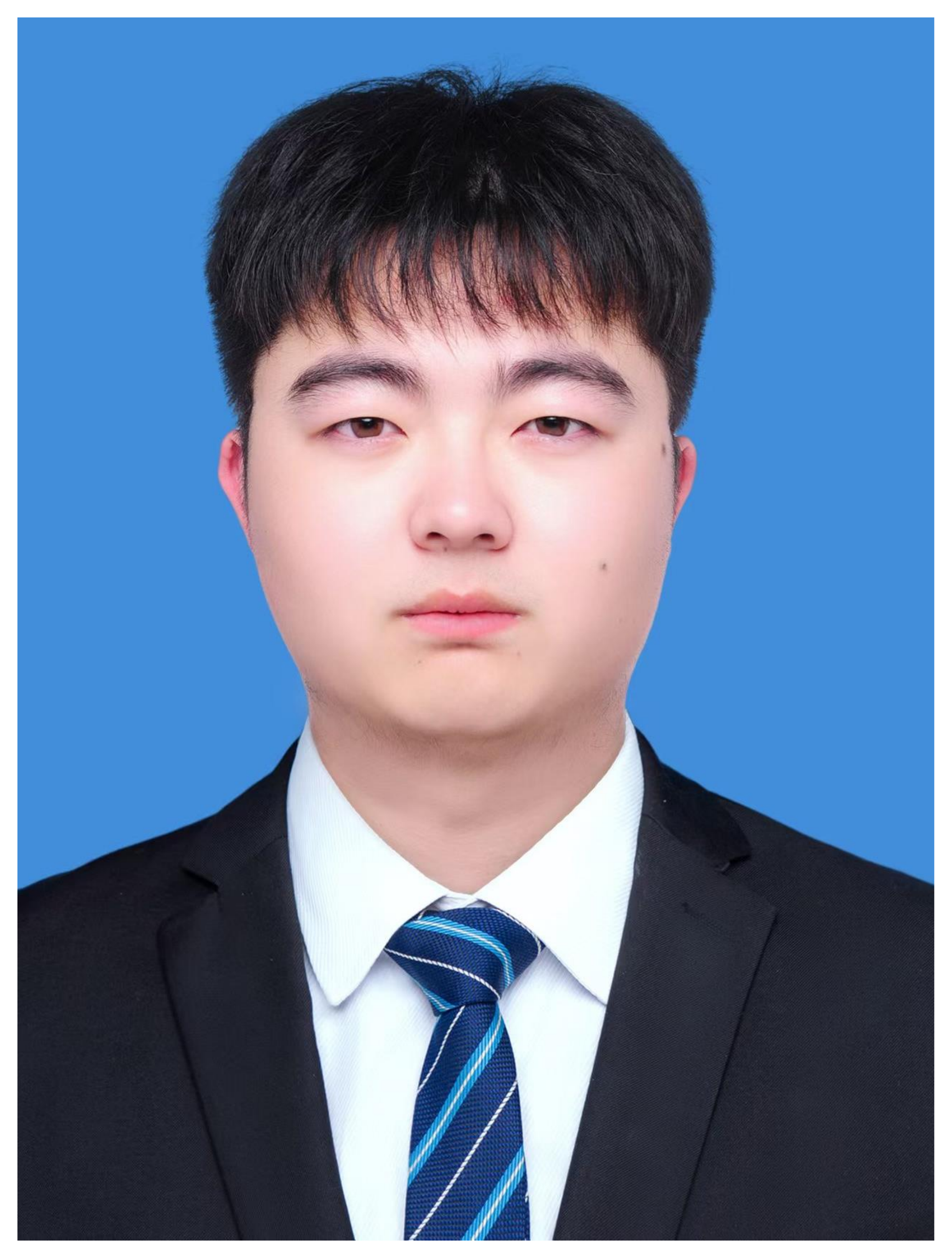}}]{Junyi Wang}
			received his bachelor’s degree from the College of Computer Science, Liaocheng University, Liaocheng, China in 2024. Currently studying for a master’s degree in the School of Computer Science and Technology, Shandong Technology and Business University, Yantai, Shandong. His research interests include computer vision and image processing.
		\end{IEEEbiography}

    \begin{IEEEbiography}[{\includegraphics[width=1in,height=1.25in,clip,keepaspectratio]{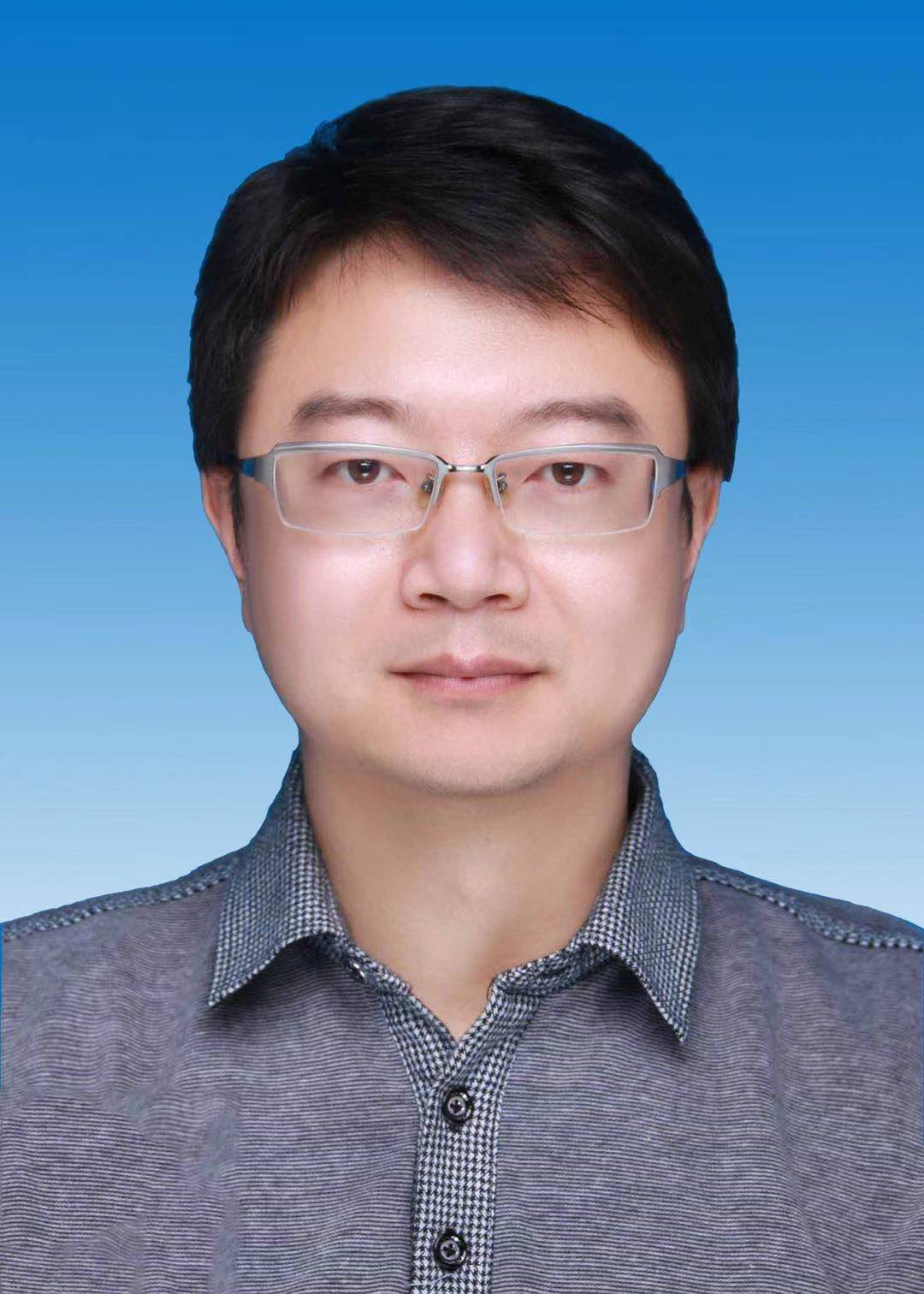}}]{Jinjiang Li}
	received the B. S. and M. S. degrees in computer science from Taiyuan University of Technology, Taiyuan, China, in 2001 and 2004, respectively, the Ph. D. degree in computer science from Shandong University, Jinan, China, in 2010. From 2004 to 2006, he was an assistant research fellow at the institute of computer science and technology of Peking University, Beijing, China. From 2012 to 2014, he was a Post-Doctoral Fellow at Tsinghua University, Beijing, China. He is currently a Professor at the school of computer science and technology, Shandong Technology and Business University. His research interests include image processing, computer graphics, computer vision, and machine learning.
\end{IEEEbiography}
\begin{IEEEbiography}[{\includegraphics[width=1in,height=1.25in,clip,keepaspectratio]{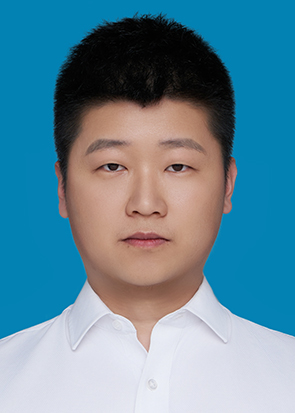}}]{Guodong Fan}
	received the M. Eng. degree from Shandong Technology and Business University, China, in 2021, the Ph.D. degree in Qingdao University, Qingdao, China, in 2025. He is currently a Associate Professor at the school of computer science and technology, Shandong Technology and Business University.  His research interests are in image processing, machine learning, and computer vision.
\end{IEEEbiography}
		\begin{IEEEbiography}
        [{\includegraphics[width=1in,height=1.25in,clip,keepaspectratio]{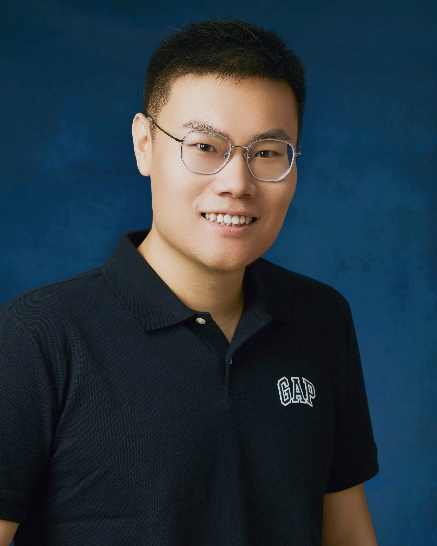}}]{Yakun Ju}  (Member, IEEE) 
 is a Lecturer (Assistant Professor) in the School of Computing and Mathematical Sciences at the University of Leicester. He previously held positions as a Research Fellow at Nanyang Technological University and a Postdoctoral Fellow at The Hong Kong Polytechnic University. He received his B.Eng. from Sichuan University in 2016 and his Ph.D. from Ocean University of China in 2022. His research focuses on 3D reconstruction, medical image processing, underwater information perception, computational imaging, and low-level vision. He has authored over 60 publications in top-tier journals and conferences, including TPAMI, IJCV, TIP, TVCG, TCSVT, CVPR, NeurIPS, AAAI, and IJCAI. He serves as an Associate Editor for Applied Soft Computing and Neurocomputing, and as a Guest Editor for Pattern Recognition and Computer Vision and Image Understanding.
\end{IEEEbiography}
\begin{IEEEbiography}[{\includegraphics[width=1.25in,height=1.25in,clip,keepaspectratio]{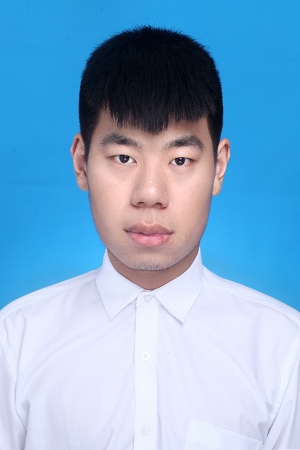}}]{Xiang Fang} received the M.E. degree from the School of Computer Science and Technology, Huazhong University of Science and Technology (HUST), in 2020. He is currently a Remote Intern with  the School of Cyber Science and Engineering, HUST. His research interests includes multi-modal learning, data mining, and machine learning.
\end{IEEEbiography}

\begin{IEEEbiography}[{\includegraphics[width=1in,height=1.25in,clip,keepaspectratio]{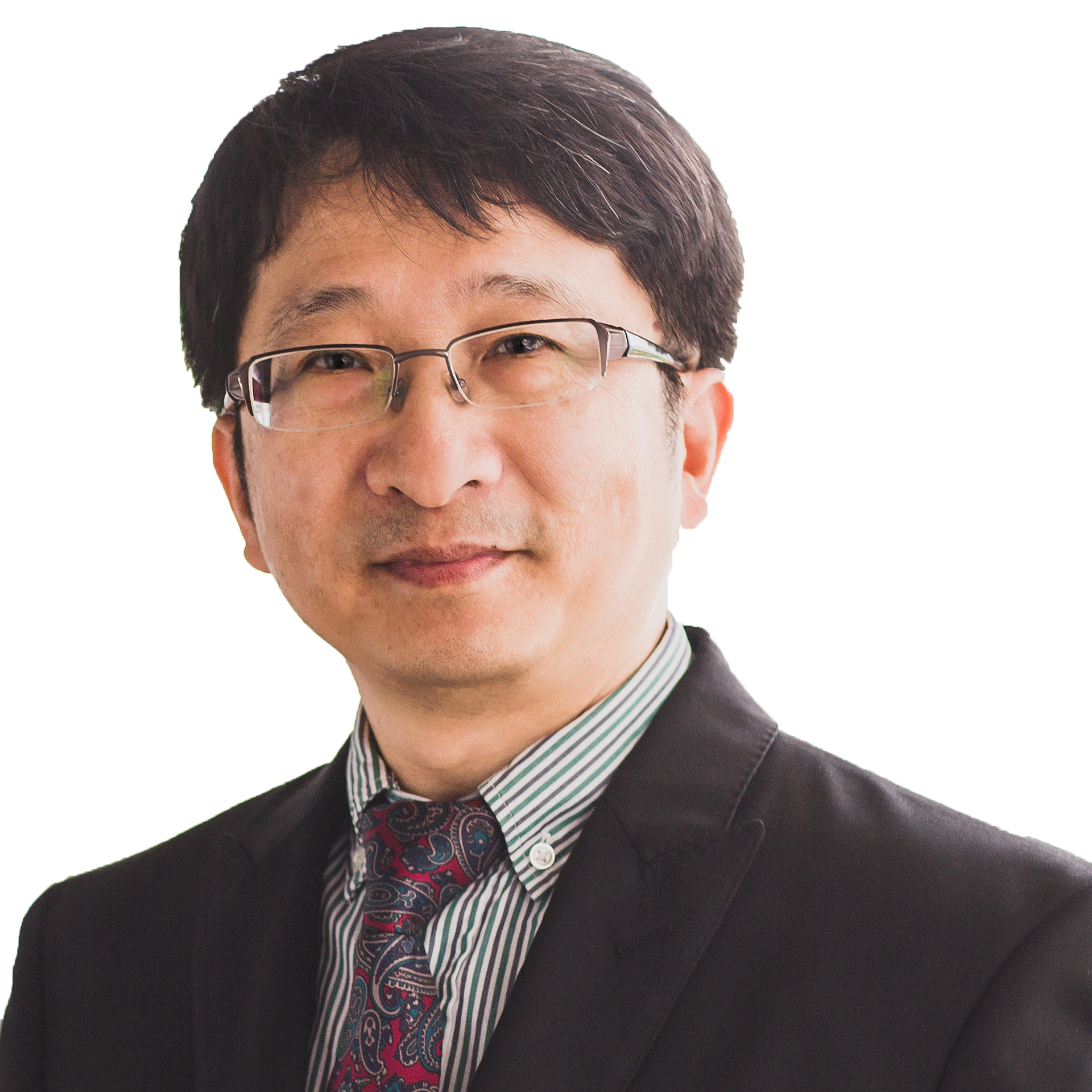}}]%
    {Alex C. Kot} (Life Fellow, IEEE)
	 has been with Nanyang Technological University, Singapore, since 1991. He was the Head of the Division of Information Engineering and the Vice Dean of Research with the School of Electrical and Electronic Engineering. Subsequently, he served as an Associate Dean for the College of Engineering for eight years. He is currently a Professor and the Director of the Rapid-Rich Object Search (ROSE) Laboratory and the NTU-PKU Joint Research Institute. He has published extensively in the areas of signal processing, biometrics, image forensics and security, and computer vision and machine learning.,He is a fellow of the Academy of Engineering, Singapore. He was elected as the IEEE Distinguished Lecturer of the Signal Processing Society and the Circuits and Systems Society. He received the Best Teacher of the Year Award. He is the co-author for several best paper awards, including ICPR, IEEE WIFS, IWDW, CVPR Precognition Workshop, and VCIP. He served at the IEEE SP Society in various capacities, such as the General Co-Chair for the 2004 IEEE International Conference on Image Processing and the Vice-President for the IEEE Signal Processing Society. He served as an associate editor for more than ten journals, mostly for IEEE Transactions.
\end{IEEEbiography}
	\end{document}